\documentclass[final,12pt]{clear2027} 

\usepackage[english]{babel}

\usepackage{tikz}
\usetikzlibrary{arrows.meta, shapes.misc}
\usepackage{standalone}
\usepackage{xparse}

\usepackage{mathtools}
\usepackage{tikz}
\usetikzlibrary{positioning}
\usepackage{enumitem}

\newtheorem{assumption}[theorem]{Assumption}

\newcommand{\pa}{pa}
\newcommand{\scm}{\mathcal{M}}
\newcommand{\scmalt}{\widetilde{\mathcal{M}}}

\newcommand{\law}[1]{P_{#1}}

\DeclareMathOperator{\doop}{do}
\newcommand{\diff}{\mathop{}\!\mathrm{d}}
\newcommand{\almev}{\text{-a.e.}}
\DeclareMathOperator{\diag}{diag}

\newcommand{\otClass}{\mathcal{C}_{\mathrm{ot}}}
\NewDocumentCommand{\triClass}{G{\pi}}{\mathcal{C}_{#1}}

\newcommand{\rvec}[1]{\mathbf{#1}}
\newcommand{\state}[1]{\mathsf{#1}}

\newcommand{\myset}[1]{\{#1\}}
\newcommand{\indep}{\perp\!\!\!\perp}

\newcommand{\Diffeo}{\mathrm{Diff}}
\DeclareMathOperator{\id}{id}
\newcommand{\R}{\mathbb{R}}
\newcommand{\N}{\mathbb{N}}
\newcommand{\norm}[1]{\left\lVert #1 \right\rVert}

\newcommand{\E}{\mathbb{E}}

\newcommand{\normal}{\mathcal{N}}
\newcommand{\bernoulli}{\mathrm{Bernoulli}}

\newcommand{\anonalt}[2]{\ifanonsubmission#1\else#2\fi}
\usepackage[nolist]{acronym}

\begin{acronym}
    \acro{scm}[SCM]{structural causal model}
    \acro{rhs}[rhs]{right hand side}
    \acro{lhs}[lhs]{left hand side}
    \acro{wlog}[w.l.o.g.]{without loss of generality}
    \acro{ci}[CI]{conditional independence}
    \acro{kr}[KR]{Knothe-Rosenblatt}
\end{acronym}

\jmlrpages{}
\jmlrproceedings{}{}

\title[Symmetries of Counterfactuals]{Symmetries of Counterfactuals: Identification and Testability of Counterfactual Maps from Covariate Heterogeneity}
\usepackage{times}

\clearauthor{%
 \Name{Marc Braun} \Email{marc.braun@liu.se}\\
 \addr Department of Computer and Information Science, Linköping University
 \AND
  \Name{Adel Daoud} \Email{adel.daoud@liu.se}\\
 \addr Institute for Analytical Sociology, Linköping University\\
 Chalmers University of Technology%
 \AND
 \Name{Jose M. Peña} \Email{jose.m.pena@liu.se}\\
 \addr Department of Computer and Information Science, Linköping University%
}

\begin{document}

\maketitle

\begin{abstract}%
In general, counterfactuals are not identified from data alone, because many causal models produce the same observed distributions but different counterfactuals. A common approach to circumvent this problem is to restrict the mechanism to a class in which it becomes unique, such as triangular monotone maps or optimal transport maps. However, whether the assumed class is correct is usually not examined. In this work, we study what the data can tell us when covariates are available that influence the outcome only through its latent cause. First, we characterise all counterfactual maps that are consistent with the data. They differ from the true map by a group of symmetries that preserve the conditional outcome distributions given the covariates. The more these distributions vary across covariate values, the smaller the group. Second, we show that for the classes above, consistency with the data is equivalent to a single conditional independence, which yields a test of the assumed class. We illustrate both results on synthetic data and on the ACTG~175 HIV trial.
\end{abstract}


\section{Introduction}

Suppose an HIV-infected patient was treated with zidovudine (ZDV) and their CD4 and CD8 cell counts were measured after 20 weeks. Would the counts have been different had the patient instead received zidovudine plus didanosine (ZDV+ddI)? A randomised trial such as ACTG~175 \citep{hammer_trial_1996} tells us how the two treatments affect the counts \emph{on average}, but a personalised recommendation requires us to answer the question for \emph{this} patient under a treatment they did not receive. Such questions are called \emph{counterfactual}. Counterfactuals play an important role in medicine, fairness \citep{kusner_counterfactual_2017}, and in explainability of algorithmic decisions \citep{karimi_algorithmic_2021}. We return to the ACTG~175 example with real data in Section~\ref{sec:actg}.

Counterfactual questions cannot be answered from data alone \citep{pearl_causality_2009}. Even if the outcome distributions under both treatments were known exactly, the data would not reveal how individual outcomes are linked across treatments. In general, infinitely many mechanisms (or \emph{outcome functions}) produce the same observed distributions and give different answers for the same patient.

A common way to circumvent this problem is to restrict the causal model to a structured class in which it becomes unique, for instance triangular monotone maps, as implemented by autoregressive normalising flows \citep{khemakhem_causal_2021}, or optimal transport maps \citep{ribeiro_counterfactual_2025}. Within such a class the counterfactual is point identified, but different classes give different answers, and whether the assumed class is correct is usually not examined. In general, it cannot be: the class is chosen precisely because, without additional information, the data does not determine the outcome function.

In this work, we ask what the data can say about the outcome function once covariates are available that are associated with the latent cause of the outcome but do not cause the outcome directly. The covariates need not be confounders. What matters is that they are informative about the latent cause, so randomised trials with informative baseline covariates are as useful as observational data. Our contributions are the following.
\begin{itemize}
    \item \textbf{Characterisation of symmetries (Section~\ref{sec:symmetry}, Theorem~\ref{thm:symmetry-group}).} The counterfactual maps consistent with the data are exactly the true map composed with a transformation that preserves the conditional outcome distributions given the covariates. These transformations form a group, the \emph{symmetry group}.
    \item \textbf{Two extremes (Propositions~\ref{pro:nasr} and~\ref{prop:no-conf-large-group}).} We show that a heterogeneity condition proposed by \citet{nasr-esfahany_counterfactual_2023-1} for continuous treatments makes the symmetry group trivial also for discrete treatments so that counterfactuals are point identified. Without any heterogeneity, every outcome is a possible counterfactual.
    \item \textbf{Testability (Section~\ref{sec:testability}, Theorem~\ref{thm:testability}).} For a candidate outcome function, a single \ac{ci} condition is equivalent to data consistency and passing the test identifies the counterfactual map up to the symmetry group. Some structural classes single out one outcome function, so testing it becomes a test about the whole structural class. We show how to measure the resolution of the test along rotations of the latent (Proposition~\ref{pro:resolution}).
    \item \textbf{Experiments (Section~\ref{sec:experiments}).} We illustrate the results on synthetic data and apply the \ac{ci} test to the ACTG~175 trial, where it rejects a reflected map decisively but cannot distinguish the triangular monotone and optimal transport classes. We quantify the resolution of the test on this data.
\end{itemize}
\section{Background and Problem Setup} \label{sec:background-and-problem-setup}

This section introduces our notation, recalls structural causal models, and states the assumptions under which our results hold.

\paragraph{Notation.}
Random variables are denoted by capital letters and their realisations by lower-case letters. We write $\law{X}$ for the distribution of a random variable $X$, i.e.\ the probability measure $B \mapsto P(X \in B)$, and $p_X$ for its density. For a measurable map $T$, the pushforward of $\law{X}$ under $T$ is $T_\#\law{X} := \law{T(X)}$.

\begin{definition}[Structural causal model]
    A \emph{\ac{scm}} is a tuple $\scm = (\rvec{X}, \rvec{U}, \mathcal{F}, P_{\rvec{U}})$ consisting of endogenous variables $\rvec{X} = (X_1, \dots, X_n)$, exogenous variables $\rvec{U} = (U_1, \dots, U_n)$ with joint distribution $P_{\rvec{U}}$, and structural functions $\mathcal{F} = \myset{f_1, \dots, f_n}$ such that $X_i = f_i\bigl((X_k)_{k \in \pa_i}, U_i\bigr)$, where $\pa_i \subseteq \myset{1, \dots, n}$ indexes the direct causes of $X_i$.
\end{definition}

An \ac{scm} induces a causal graph with nodes $X_1, \dots, X_n$ and an edge $X_k \to X_i$ whenever $k \in \pa_i$, and a distribution $P^{\scm}$ over $\rvec{X}$. The exogenous variables are unobserved. Two \acp{scm} over the same endogenous variables are \emph{observationally equivalent} if they induce the same $P^{\scm}$. An intervention $\doop(X_i = x_i)$ replaces $f_i$ by the constant $x_i$, and we write $\scm_{\doop(X_i = x_i)}$ for the resulting \ac{scm} and $P^{\scm}(\rvec{X} \mid \doop(X_i = x_i)) \coloneq P^{\scm_{\doop(X_i = x_i)}}(\rvec{X})$ for the interventional distribution.

\paragraph{Problem setup.}
Let $\scm = ( (V, A, Y), (U_V, U_A, U), \mathcal{F}, P_{( U_V, U_A, U )})$ be an \ac{scm} where $A$ takes values in a finite set $\state{A}$, $Y$ takes values in $\R^d$, and $V$ takes values in $\R^{d_V}$. Let $(f_a)_{a \in \state{A}}$ be a family of functions with $f_a: \R^d \rightarrow \R^d$.

\begin{assumption}
    We make the following structural assumptions about $\scm$.
    \begin{enumerate}[label=(\roman*)]\label{ass:structural-assumptions}
        \item $Y = f_A(U)$ \label{ass:outcome-func-of-u-a}
        \item $U \indep A \mid V$ \label{ass:u-indep-a-given-v}
        \item $f_a \in \Diffeo^1(\R^d)$ for all $a \in \state{A}$, i.e. $f_a$ is a bijection with $f_a$ and $f_a^{-1}$ continuously differentiable \label{ass:y-is-diffeo}
        \item Overlap: $P(A=a \mid V=v) > 0$ for all $a \in \state{A}$ and $\law{V} \almev\ v$ \label{ass:overlap}
    \end{enumerate}
\end{assumption}

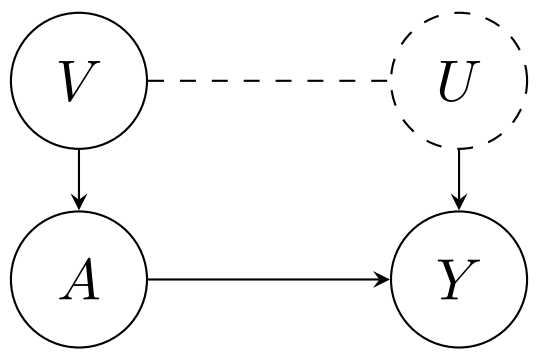
\begin{figure}[ht]
    \begin{center}
    \begin{tikzpicture}[
        node distance=0.4cm and 1.6cm,
        endogenous/.style={draw, circle, minimum size=0.9cm},
        exogenous/.style={draw, circle, dashed, minimum size=0.9cm},
        >=stealth
    ]
        \node[endogenous] (V) {$V$};
        \node[endogenous] (A) [below=of V] {$A$};
        \node[endogenous] (Y) [right=of A] {$Y$};
    
        \node[exogenous] (U) [above=of Y] {$U$};
    
    
        \draw[->] (V) -- (A);
        \draw[->] (A) -- (Y);
        \draw[-, dashed] (V) -- (U);
        \draw[->] (U) -- (Y);
    \end{tikzpicture}
    \end{center}
    \caption{Causal graph of our setup. The latent variable $U$ is drawn as a dashed node. The dashed edge between $U$ and $V$ denotes that they may be dependent.}
    \label{fig:causal-graph}
\end{figure}

The causal graph of this \ac{scm} is shown in Figure~\ref{fig:causal-graph}.

The variable $Y$ represents an outcome variable of interest and the variable $A$ some treatment that effects $Y$ through the function $f_A$. We introduce the following shorthand notation for the distributions we use most often: $\mu \coloneq \law{U}, \mu_v \coloneq \law{U \mid V=v}, \nu_{a, v} \coloneq \law{Y \mid A=a, V=v}, \nu_a \coloneq \law{Y \mid \doop(A=a)}$.
The variable $V$ is sufficient to control for confounding between $A$ and $Y$, because $\nu_a = {(f_a)}_\# \mu = \int \nu_{a, v} \diff \law{V}(v)$. The first equality follows from the definition of the interventional distribution and Assumption~\ref{ass:structural-assumptions}~\ref{ass:outcome-func-of-u-a}, and the second one from marginalising over $V$ and using Assumptions~\ref{ass:structural-assumptions}~\ref{ass:u-indep-a-given-v} and~\ref{ass:overlap}. We therefore may write $\nu_{a,v}$ and $\law{Y \mid \doop(A=a), V=v}$ interchangeably.

\begin{definition}[Data consistency]
    We say that an \ac{scm} is \emph{data consistent} if it is observationally equivalent to the true \ac{scm} $\scm$, i.e. its induced data distribution is equal to the observed data distribution $\law{V, A, Y}$, and if it satisfies Assumption~\ref{ass:structural-assumptions}.
\end{definition}

\begin{definition}[Counterfactual map and outcome]\label{def:counterfactual-map-and-outcome}
    We call
    $T^{f}_{a \rightarrow a'} \coloneq f_{a'} \circ f_a^{-1}$
    the \emph{counterfactual map} from $a$ to $a'$ for outcome function $f$. Given an observation $(a, y)$, we call $T^{f}_{a \rightarrow a'}(y)$ the \emph{counterfactual outcome} of $y$ under $\doop(A=a')$.
\end{definition}

In our setup, this definition corresponds exactly to the classical abduction, action, prediction framework for counterfactual inference proposed by \citet{pearl_causality_2009}.

Given an observation $(v, a, y)$, practical applications often require us to answer the counterfactual question what the outcome $y$ would have been had the treatment been $a' \neq a$. After reviewing related work in the next section, we will characterise the set of all counterfactual maps that are data consistent in Section~\ref{sec:symmetry}.
\section{Related Work}\label{sec:related-work}

\paragraph{Non-identifiability and partial identification.}
For a one-dimensional outcome and a strictly monotone outcome function, counterfactuals are identified: the latent is the rank of the outcome, and the counterfactual map is the quantile-to-quantile map between treatment arms \citep{lu_sample-efficient_2020, nasr-esfahany_counterfactual_2023}. Without monotonicity, counterfactuals are not identified even in one dimension, and \citet{melnychuk_partial_2023} derive bounds that depend on the curvature of the outcome function. In more than one dimension, bijectivity alone does not help: many \acp{scm} with bijective outcome functions induce the same observed distribution but different counterfactuals \citep{nasr-esfahany_counterfactual_2023}. Our Theorem~\ref{thm:symmetry-group} describes this set of \acp{scm} exactly, and Proposition~\ref{prop:no-conf-large-group} shows that without covariate heterogeneity it is as large as it can be.

\paragraph{Point identification through structured classes.}
To restore point identification, the outcome function is restricted to a class in which it is unique. \citet{de_lara_transport-based_2024} use optimal transport maps between the interventional distributions as counterfactual maps. Such maps are unique for a given cost, but they do not compose: transporting from $a_0$ to $a_1$ and then to $a_2$ need not agree with transporting from $a_0$ to $a_2$ directly. \citet{ribeiro_counterfactual_2025} require the outcome functions $f_a$ themselves to be optimal transport maps from a fixed latent reference distribution $\eta$, which yields transitive counterfactual maps but, as we show in Appendix~\ref{app:ot-counterexample}, makes them depend on the choice of $\eta$. The other common class consists of triangular monotone outcome functions, which are unique for any reference distribution and are implemented by autoregressive normalising flows \citep{khemakhem_causal_2021, javaloy_causal_2023, chen_exogenous_2025}. Deep structural causal models \citep{pawlowski_deep_2020, balgi_deep_2025} learn outcome functions in such classes from data. Closest to our work, \citet{nasr-esfahany_counterfactual_2023-1} give a sufficient condition for point identification of bijective outcome functions when covariates of the outcome are available and the treatment is continuous. We adapt it to discrete treatments in Proposition~\ref{pro:nasr} and characterise what remains unidentified when it fails.

\paragraph{Nonlinear ICA.}
Our setting is related to nonlinear ICA with auxiliary variables. There, a latent variable is observed through an unknown invertible transformation, which is not identifiable from the observed distribution alone \citep{hyvarinen_nonlinear_1999}. Identifiability up to a restricted class of transformations is recovered when an auxiliary variable modulates the latent distribution sufficiently \citep{hyvarinen_nonlinear_2019, khemakhem_variational_2020}. In our setup the covariate $V$ plays the role of the auxiliary variable, and the condition of Proposition~\ref{pro:nasr} is an analogue of the sufficient-variability conditions in that literature. Our target differs: we are interested in the counterfactual map $T^f_{a \to a'}$ rather than in the latent itself, and we use the auxiliary variable not only for identification but to test a hypothesised model class.
\section{Symmetry Group for Counterfactuals}\label{sec:symmetry}

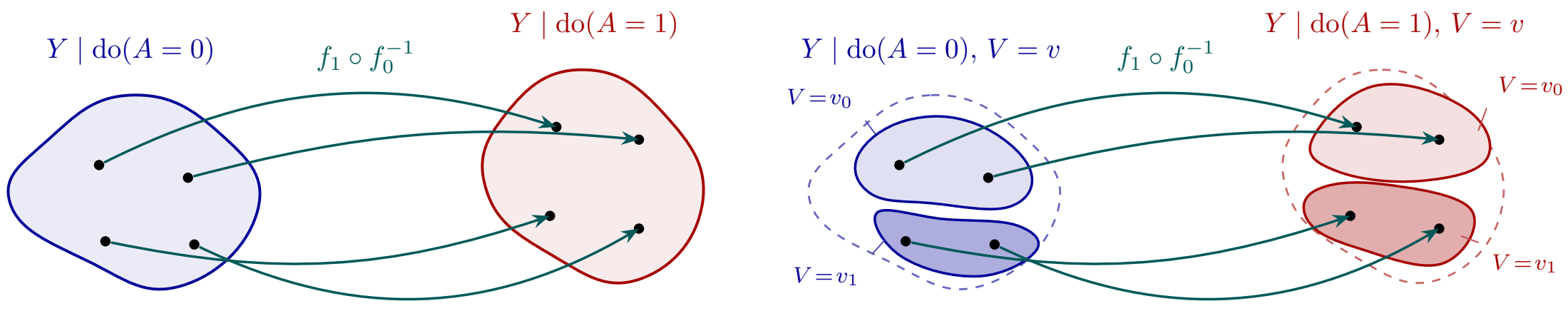
\begin{figure}[t]
  \centering
  \resizebox{\linewidth}{!}{
\providecommand{\doop}{\operatorname{do}}

\def\leftblob{plot [smooth cycle, tension=0.8] coordinates
  {(0,0.2) (1.3,0.9) (2.7,0.3) (3.3,-0.9) (2.1,-2.1) (0.6,-1.7) (-0.6,-0.8)}}
\def\rightblob{plot [smooth cycle, tension=0.8] coordinates
  {(7.2,0.6) (8.4,1.3) (9.7,0.6) (10.3,-0.8) (9.3,-2.0) (7.9,-1.6) (6.9,-0.6)}}

\def\leftvzero{plot [smooth cycle, tension=0.7] coordinates
  {(0.15,-0.05) (0.9,0.5) (2.0,0.45) (2.85,-0.25) (2.55,-0.85) (1.3,-0.8) (0.35,-0.65)}}
\def\rightvzero{plot [smooth cycle, tension=0.7] coordinates
  {(7.3,0.2) (8.3,1.05) (9.7,0.7) (10.05,-0.1) (9.0,-0.45) (7.7,-0.3)}}

\def\leftvone{plot [smooth cycle, tension=0.7] coordinates
  {(0.45,-0.95) (1.5,-1.05) (2.65,-1.15) (2.95,-1.55) (2.05,-1.95) (0.85,-1.65)}}
\def\rightvone{plot [smooth cycle, tension=0.7] coordinates
  {(7.35,-0.6) (8.4,-0.5) (9.8,-0.9) (9.35,-1.75) (8.0,-1.5) (7.25,-1.1)}}

\newcommand{\points}{%
  \node[pt] (a1) at (0.8,-0.2)  {};
  \node[pt] (a2) at (2.2,-0.4)  {};
  \node[pt] (a3) at (0.9,-1.4)  {};
  \node[pt] (a4) at (2.3,-1.45) {};
  \node[pt] (b1) at (8.0,0.4)   {};
  \node[pt] (b2) at (9.3,0.2)   {};
  \node[pt] (b3) at (7.9,-1.0)  {};
  \node[pt] (b4) at (9.3,-1.2)  {};
}
\newcommand{\arrows}{%
  \draw[map] (a1) to[bend left=22]  (b1);
  \draw[map] (a2) to[bend left=10]  (b2);
  \draw[map] (a3) to[bend right=15] (b3);
  \draw[map] (a4) to[bend right=28] (b4);
}

\begin{tikzpicture}[
  >={Stealth[length=2.2mm,width=1.7mm]},
  blob/.style={draw=#1, fill=#1!8, line width=1pt},
  ghost/.style={draw=#1!60, dashed, line width=0.7pt},
  condA/.style={draw=#1, fill=#1!12, line width=0.9pt},   
  condB/.style={draw=#1, fill=#1!32, line width=0.9pt},   
  pt/.style={circle, fill=black, inner sep=1.3pt},
  map/.style={->, draw=teal!70!black, line width=0.9pt},
  lab/.style={font=\small},
  panel/.style={font=\small},
  vlab/.style={font=\scriptsize},
  leader/.style={draw=#1!70, line width=0.5pt, shorten >=1pt, shorten <=1pt},
  scale=0.8,
]

\begin{scope}[name prefix=m-]
  \node[panel] at (4.85,-3.1) {\textbf{(a)} marginal};
  \draw[blob=blue!60!black] \leftblob;
  \draw[blob=red!65!black]  \rightblob;
  \node[lab, blue!60!black] at (1.3,1.6) {$Y \mid \doop(A=0)$};
  \node[lab, red!65!black]  at (8.6,2.0) {$Y \mid \doop(A=1)$};
  \points
  \arrows
  \node[lab, teal!70!black] at (5.0,1.5) {$f_1 \circ f_0^{-1}$};
\end{scope}

\begin{scope}[name prefix=c-, shift={(12.6,0)}]
  \node[panel] at (4.85,-3.1) {\textbf{(b)} conditional on $V$};
  \draw[ghost=blue!60!black] \leftblob;
  \draw[ghost=red!65!black]  \rightblob;
  \draw[condA=blue!60!black] \leftvzero;
  \draw[condB=blue!60!black] \leftvone;
  \draw[condA=red!65!black]  \rightvzero;
  \draw[condB=red!65!black]  \rightvone;
  \node[lab, blue!60!black] at (1.3,1.6) {$Y \mid \doop(A=0),\,V=v$};
  \node[lab, red!65!black]  at (8.6,2.0) {$Y \mid \doop(A=1),\,V=v$};
  \node[vlab, blue!60!black] (lv0) at (-0.45,0.85)  {$V\!=\!v_0$};
  \node[vlab, blue!60!black] (lv1) at (-0.35,-1.95) {$V\!=\!v_1$};
  \node[vlab, red!65!black]  (rv0) at (10.75,1.05)  {$V\!=\!v_0$};
  \node[vlab, red!65!black]  (rv1) at (10.65,-1.75) {$V\!=\!v_1$};
  \draw[leader=blue!60!black] (lv0.south east) -- (0.45,0.25);
  \draw[leader=blue!60!black] (lv1.north east) -- (0.6,-1.35);
  \draw[leader=red!65!black]  (rv0.south west) -- (9.9,0.3);
  \draw[leader=red!65!black]  (rv1.north west) -- (9.6,-1.3);
  \points
  \arrows
  \node[lab, teal!70!black] at (5.0,1.5) {$f_1 \circ f_0^{-1}$};
\end{scope}

\end{tikzpicture}}
  \caption{The transport map $T^f_{0 \to 1} = f_1 \circ f_0^{-1}$ pushes $Y \mid \doop(A{=}0)$ forward to $Y \mid \doop(A{=}1)$ (a). Conditioning on $V$ does not change the map, but it must now also push each conditional distribution $Y \mid \doop(A{=}0), V{=}v$ to $Y \mid \doop(A{=}1), V{=}v$ for every $v$ (b, shaded regions). The more these conditional distributions vary with $v$, the fewer maps satisfy the constraint.}
  \label{fig:intuition}
\end{figure}

Before we formally state our characterisation of the set of data consistent counterfactual maps, we provide some intuition.
Let us assume for now that $A$ is binary and we would like to know the counterfactual map from $A=0$ to $A=1$. Then we know that this map has to transport the interventional distribution of $Y \mid \doop(A=0)$ to the distribution of $Y \mid \doop(A=1)$. We call this the \emph{marginal transport} condition and illustrate it in Figure~\ref{fig:intuition} (a). In more than one dimension, there are infinitely many diffeomorphisms that can realise this transport. For example, when $\nu_0$ is a standard Gaussian, then $f_1 \circ f_0^{-1} \circ R$ where $R$ is a rotation is another such function.
However, according to our setup, the counterfactual map also has to transport the conditional distribution of $Y \mid \doop(A=0), V=v$ to the distribution of $Y \mid \doop(A=1), V=v$ for all $v$. We call this the \emph{conditional transport} condition (Figure~\ref{fig:intuition} (b)).
At the same time, the counterfactual map $T^f_{0 \to 1} = f_1 \circ f_0^{-1}$ does not change with $v$ because $V$ is not a direct cause of $Y$ according to Assumption~\ref{ass:structural-assumptions}~\ref{ass:outcome-func-of-u-a}. Therefore, every new conditional distribution $\nu_{a, v}$ adds another restriction, so heterogeneity of $\nu_{a, v}$ with respect to $v$ helps to find the correct counterfactual map. This can be seen in Figure~\ref{fig:intuition} (b), where for example the bottom two points cannot be moved to any of the top two points, because $Y \mid \doop(A=1), V=v_1$ has no support there. In general the conditional transport condition introduces additional constraints even when the supports overlap.

We characterise the exact set of counterfactual maps that are consistent with the data in the following theorem.

\begin{theorem}[Counterfactual Symmetry Group]\label{thm:symmetry-group}
    Let $a, a' \in \state{A}$ with $a \neq a'$ and let $G_{a} \coloneq \myset{M \in \Diffeo^1(\R^d) : M_\# \nu_{a,v} = \nu_{a, v} \text{ for } \law{V}\almev\ v}$.
    The set of counterfactual maps $a \to a'$ achievable by data consistent \acp{scm} is exactly $\myset{T^{f}_{a \to a'} \circ M : M \in G_{a}}$.
    Furthermore, for all $a \in \state{A}$ the tuple $(G_{a}, \circ)$ with the composition operator $\circ$ is a group.
\end{theorem}

Theorem~\ref{thm:symmetry-group} tells us that the counterfactual maps from $a$ to $a'$ that are consistent with the data differ exactly by those transformations $M \in G_{a}$ that preserve the conditional distribution $\nu_{a, v}$ for $\law{V}\almev\ v$. It furthermore allows us to characterise the set of counterfactual outcomes for an observation $(y, a)$ under intervention $\doop(A=a')$ that can be reached by data consistent \acp{scm} as $T^f_{a \to a'}(G_a \cdot y)$ where $G_a \cdot y \coloneq \myset{M(y) : M \in G_a}$ is the orbit of $y$ under $G_a$. When $a$, $a'$, and $y$ are clear from the context, we call $T^f_{a \to a'}(G_a \cdot y)$ the \emph{identified set} of counterfactuals. Counterfactuals are point identified if $T^f_{a \to a'}(G_a \cdot y) = \myset{T^f_{a \to a'}(y)}$ for all $y$. In particular, counterfactuals $a \to a'$ are point identified if and only if $G_a = \myset{\id}$ since $T^f_{a \to a'}$ is a bijection.

Proposition~\ref{pro:nasr} shows that a condition proposed by \citet{nasr-esfahany_counterfactual_2023-1}  for continuous treatments also yields point identification for discrete treatments, and connects it to the symmetry group $G_a$.

\begin{proposition}[Point identification]\label{pro:nasr}
    Let $p_v(u) \coloneq p_{U \mid V=v}(u)$ be continuously differentiable. If there exist $d+1$ values $v_0, \dots v_d$ with $P(V=v_j) > 0$ such that
    \[
    \left| \det \begin{pmatrix}
        p_{v_0}(u) & D p_{v_0}(u) \\
        \vdots & \vdots \\
        p_{v_d}(u) & D p_{v_d}(u)
    \end{pmatrix} \right|
    > 0 \qquad \text{for all } u,
    \]
    then $G_a$ is trivial for all $a$, i.e. $G_a = \myset{\id}$.
\end{proposition}

The condition requires $d+1$ sufficiently different conditional distributions of $U$ and is sufficient for point identification. Section~\ref{sec:toy-example} illustrates how the symmetry group shrinks as their number grows.
The opposite extreme is when there is no heterogeneity at all, as stated in Proposition~\ref{prop:no-conf-large-group}.

\begin{proposition}[No heterogeneity]\label{prop:no-conf-large-group}
Let $d \ge 2$ and let $\nu_a$ have a smooth, everywhere positive density. If $U \indep V$ (equivalently, $\nu_{a,v} = \nu_a$ for $\law{V}\almev\ v$), then for all $a, a' \in \state{A}$ and every $y \in \R^d$,
\[
  T^f_{a \to a'}(G_a \cdot y) = \R^d.
\]
\end{proposition}

What we can conclude from Proposition~\ref{pro:nasr} and from Proposition~\ref{prop:no-conf-large-group} is that enough heterogeneity in the data reduces the symmetry group to $\myset{\id}$, while independence of $U$ and $V$ makes the group so large that counterfactuals are fundamentally unidentified. Loosely speaking, we can say that heterogeneity helps identifiability while no heterogeneity destroys it.
\section{Testability of Structural Assumptions}\label{sec:testability}

Theorem~\ref{thm:symmetry-group} describes what the data can say about counterfactual maps in general. In practice, one usually assumes a structured class of outcome functions, such as triangular monotone or optimal transport maps, for which the marginal transport condition already determines the outcome function (Figure~\ref{fig:intuition}~(a)). We show that the conditional transport condition (Figure~\ref{fig:intuition}~(b)) makes the assumption testable: a hypothesised outcome function is compatible with the data if and only if a single \ac{ci} holds, and the more heterogeneous the conditional distributions are, the more this test can tell us.

\begin{definition}[Function and model classes]
    Let
    \begin{align*}
    \triClass
    &\coloneq
    \myset{g \in \Diffeo^1(\R^d) \text{, triangular monotone in ordering } \pi}, \\
    \otClass
    &\coloneq
    \myset{g \in \Diffeo^1(\R^d) \text{ with } g = \nabla \phi \text{ for a convex } \phi : \R^d \to \R}.
    \end{align*}
    For a function class $\mathcal{C}$ and a distribution $\eta$ on $\R^d$, the \emph{model class} $\scm(\mathcal{C}, \eta)$ is the set of \acp{scm} that satisfy Assumption~\ref{ass:structural-assumptions} with $f_a \in \mathcal{C}$ for all $a \in \state{A}$ and $\law{U} = \eta$.
\end{definition}

We write $f \in \mathcal{C}^{\state{A}}$ if $f_a \in \mathcal{C}$ for all $a \in \state{A}$. Triangular monotone maps are what autoregressive normalising flows implement, and gradients of convex functions are the optimal transport maps for the quadratic cost.

\begin{definition}[Selective class]\label{def:selective-class}
    A model class $\scm(\mathcal{C}, \eta)$ is called \emph{selective} if for every $a \in \state{A}$ there is at most one\footnote{We require uniqueness only \emph{if} a canonical map exists: the unique transport map need not be a diffeomorphism, in which case it violates Assumption~\ref{ass:structural-assumptions}~\ref{ass:y-is-diffeo} and the class contains no data-consistent \ac{scm}.} (up to $\eta \almev$ equality) $h_a \in \mathcal{C}$ with ${h_a}_\# \eta = \nu_a$. If such maps exist for all $a$, we write $h^{\mathcal{C}}_a$ for them and call $h^{\mathcal{C}}$ the \emph{canonical outcome function} and $T^{h^{\mathcal{C}}}_{a \to a'}$ the \emph{canonical counterfactual map}.
\end{definition}

For any absolutely continuous $\eta$, both $\scm(\triClass, \eta)$ and $\scm(\otClass, \eta)$ are selective (Appendix~\ref{app:tri-ot-selective}), so the marginal transport alone, the only constraint available without covariates, already singles out one outcome function. For $\triClass$ the canonical outcome function is called the \ac{kr} map, and for $\otClass$ the Brenier map. The two classes differ in how they depend on the choice of $\eta$. For $\triClass$ the canonical counterfactual map does not depend on $\eta$ (Appendix~\ref{app:tri-ot-selective}), so the reference distribution can be chosen freely. For $\otClass$ it does in general depend on $\eta$ (Appendix~\ref{app:ot-counterexample}), so the reference distribution is part of the hypothesis being tested.

\begin{remark}\label{rem:ot-needs-fixed-reference}
    The reason for this difference is that the proof for $\triClass$ (Appendix~\ref{app:tri-ot-selective}) relies on closure under composition of triangular monotone maps. The composition of two gradients of convex functions is in general not a gradient of a convex function, so the argument does not carry over to $\otClass$, and the counterexample in Appendix~\ref{app:ot-counterexample} shows that uniqueness within $\scm(\otClass, \eta)$ for a fixed $\eta$, as in Theorem~4.12 of \citet{ribeiro_counterfactual_2025}, does not extend across reference distributions.
\end{remark}

The canonical outcome function of a selective class is the only candidate for a data-consistent \ac{scm} within the class, because it is the only one that satisfies the marginal transport condition (Appendix~\ref{app:selective-consequences}). This follows from the definition of a selective class, but the definition says nothing about the conditional transport condition, which is what data consistency requires. Theorem~\ref{thm:symmetry-group} characterises the set of data-consistent counterfactual maps, but it does not tell us how to check whether a given outcome function $h$ belongs to it. Theorem~\ref{thm:testability} provides such a check in the form of a single \ac{ci} condition.

\begin{theorem}[Testability]\label{thm:testability}
Let $h_a \in \Diffeo^1(\R^d)$ for all $a \in \state{A}$. Then the following are equivalent:
  \begin{enumerate}
  \item[(i)] there exists a data-consistent \ac{scm} with outcome function $h$,
  \item[(ii)] $h_A^{-1}(Y) \indep A \mid V$,
  \item[(iii)] $T^{h}_{a \to a'} \in \myset{T^f_{a \to a'} \circ M : M \in G_a}$ for all $a, a' \in \state{A}$.
  \end{enumerate}
\end{theorem}

\begin{figure}[t]
  \centering
  \resizebox{\linewidth}{!}{\providecommand{\doop}{\operatorname{do}}

\def\leftblob{plot [smooth cycle, tension=0.8] coordinates
  {(0,0.2) (1.3,0.9) (2.7,0.3) (3.3,-0.9) (2.1,-2.1) (0.6,-1.7) (-0.6,-0.8)}}
\def\rightblob{plot [smooth cycle, tension=0.8] coordinates
  {(7.2,0.6) (8.4,1.3) (9.7,0.6) (10.3,-0.8) (9.3,-2.0) (7.9,-1.6) (6.9,-0.6)}}

\newcommand{\arms}{%
  \draw[cond=blue!60!black] \leftblob;
  \node[lab, blue!60!black] at (1.3,1.55) {$Y \mid A=0,\,V=v$};
  \begin{scope}[shift={(2.3,0)}]
    \draw[cond=red!65!black] \rightblob;
    \node[lab, red!65!black] at (8.6,1.85) {$Y \mid A=1,\,V=v$};
  \end{scope}
  \node[pt] (a1) at (0.6,0.1)   {};
  \node[pt] (a2) at (2.3,-0.1)  {};
  \node[pt] (a3) at (1.5,-1.3)  {};
  \node[cr] (b1) at (10.3,0.4)  {};
  \node[cr] (b2) at (11.6,0.2)  {};
  \node[cr] (b3) at (10.8,-1.1) {};
}

\begin{tikzpicture}[
  >={Stealth[length=2.0mm,width=1.5mm]},
  cond/.style={draw=#1, fill=#1!12, line width=0.9pt},
  latent/.style={draw=#1, fill=#1!12, line width=0.9pt},
  pulled/.style={draw=#1, densely dashed, line width=1.0pt},
  pt/.style={circle, fill=black, inner sep=1.2pt},
  cr/.style={cross out, draw=black, line width=0.9pt, minimum size=4.5pt, inner sep=0pt},
  mapz/.style={->, draw=blue!60!black, line width=0.9pt},
  mapo/.style={->, draw=red!65!black, line width=0.9pt},
  lab/.style={font=\footnotesize},
  panel/.style={font=\footnotesize},
  scale=0.6,
]

\begin{scope}[name prefix=a-]
  \arms
  \begin{scope}[shift={(4.6,0)}]
    \draw[latent=gray!70!black] \leftblob;
    \node[lab, gray!40!black] at (1.35,1.55) {$U \mid V=v$};
    \node[pt] (u1) at (0.6,0.1)  {};
    \node[pt] (u2) at (2.3,-0.1) {};
    \node[pt] (u3) at (1.5,-1.3) {};
  \end{scope}
  \draw[mapz] (u1) to[bend right=18] (a1);
  \draw[mapz] (u2) to[bend right=10] (a2);
  \draw[mapz] (u3) to[bend left=12]  (a3);
  \draw[mapo] (u1) to[bend left=18]  (b1);
  \draw[mapo] (u2) to[bend left=10]  (b2);
  \draw[mapo] (u3) to[bend right=12] (b3);
  \node[lab, blue!60!black]  at (3.65,-2.25) {$f_0$};
  \node[lab, red!65!black]   at (8.55,-2.25) {$f_1$};
  \node[panel] at (6.0,-3.1)
    {\textbf{(a)} Both conditionals come from one latent distribution.};
\end{scope}

\begin{scope}[name prefix=b-, shift={(14.4,0)}]
  \arms
  \begin{scope}[shift={(4.6,0)}]
    \draw[pulled=blue!60!black] \leftblob;
    \node[lab, gray!40!black] at (1.35,1.55) {$h_A^{-1}(Y) \mid V=v$};
    \node[pt] (c1) at (0.45,0.2)   {};
    \node[pt] (c2) at (2.15,-0.0)  {};
    \node[pt] (c3) at (1.35,-1.2)  {};
    \begin{scope}[shift={(0.35,0.25)}, rotate around={-30:(1.4,-0.75)}]
      \draw[pulled=red!65!black] \leftblob;
      \node[cr] (x1) at (0.75,-0.0)  {};
      \node[cr] (x2) at (2.45,-0.2)  {};
      \node[cr] (x3) at (1.65,-1.4)  {};
    \end{scope}
  \end{scope}
  \draw[mapz] (a1) to[bend left=18]  (c1);
  \draw[mapz] (a2) to[bend left=10]  (c2);
  \draw[mapz] (a3) to[bend right=12] (c3);
  \draw[mapo] (b1) to[bend right=18] (x1);
  \draw[mapo] (b2) to[bend right=10] (x2);
  \draw[mapo] (b3) to[bend left=12]  (x3);
  \node[lab, blue!60!black]  at (3.65,-2.25) {$h_0^{-1}$};
  \node[lab, red!65!black]   at (8.55,-2.25) {$h_1^{-1}$};
  \node[panel] at (6.0,-3.1)
    {\textbf{(b)} Pulled back, the conditionals must coincide.};
\end{scope}

\end{tikzpicture}}
  \caption{Intuition for (ii) of Theorem~\ref{thm:testability}. (a) Since $U \indep A \mid V$, both conditionals $Y \mid A = a, V = v$ are images of the same latent distribution. (b) Pulled back with $h$, they must therefore coincide. Requiring this for every $v$ is the independence of Theorem~\ref{thm:testability}~(ii). In the example in (b) they do not coincide, so an eventual hypothesis test will likely be rejected.}
  \label{fig:testability-intuition}
\end{figure}
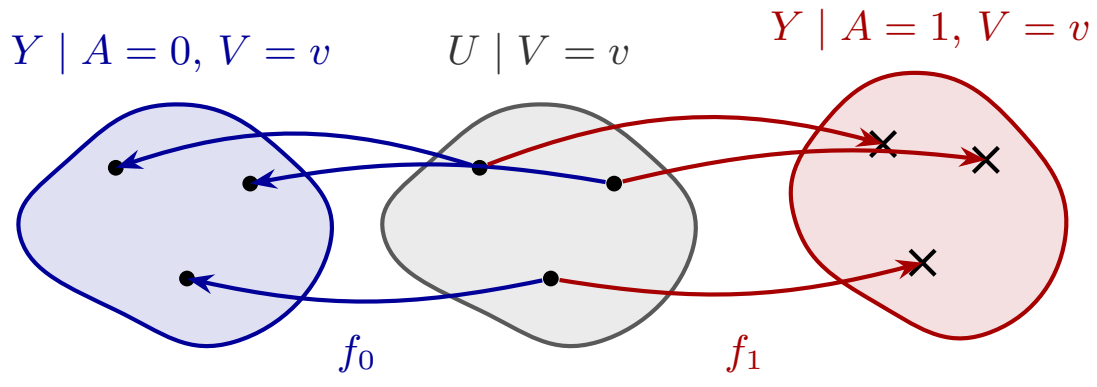

The \ac{ci} requirement (ii) in Theorem~\ref{thm:testability} is a testable form of the conditional transport condition from Section~\ref{sec:symmetry}: for binary $A$ it holds if and only if $T^h_{0 \to 1}$ transports $\nu_{0, v}$ to $\nu_{1, v}$ for $\law{V} \almev\ v$. Testing whether the distribution induced by the outcome function $h$ is equal to the observed distribution $\law{V, A, Y}$ directly is not possible, because the distribution of the latent conditioned on $V$ is unspecified for a model class $\scm(\mathcal{C}, \eta)$ even when $\eta$ is fixed. Condition (ii) is this comparison with the latent eliminated: both arms are pulled back to the latent space where they must agree (Figure~\ref{fig:testability-intuition}).

Theorem~\ref{thm:testability} tells us when an outcome function $h$ is consistent with the data, not whether its counterfactual map $T^h_{a \to a'}$ is the true one. If the \ac{ci} test rejects, then the counterfactual map differs from the true one. If it does not reject, then the counterfactual map $T^h_{a \to a'}$ lies in $\myset{T^{f}_{a \rightarrow a'} \circ M : M \in G_a}$ and may still differ from the true map by an element of $G_a$. The more the conditional distributions vary across covariate values, the smaller the group $G_a$ and the more wrong counterfactual maps the test can detect. When $G_a$ is trivial, every wrong map is detectable and passing the test identifies the true counterfactual map. In the opposite case where $U$ is independent of $V$, the canonical outcome function of any selective class is the outcome function of a data-consistent \ac{scm}, because the conditional transport condition then reduces to the marginal one (Appendix~\ref{app:selective-consequences}). Without heterogeneity, structural assumptions are therefore not testable.

In practice, a \ac{ci} test on limited data may lack the power to reject an outcome function that is not data consistent, so a non-rejection could be either due to low power or because the outcome function is indeed data consistent. The following proposition lets us tell the two apart along rotations.

\begin{proposition}[Resolution along rotations]\label{pro:resolution}
Let $d = 2$, $h_a \in \Diffeo^1(\R^2)$ for all $a \in \state{A}$, and fix $a \neq a'$. For a rotation $R_\theta$ by angle $\theta$ define $h^\theta$ by $h^\theta_{a'} \coloneq h_{a'} \circ R_\theta$ and $h^\theta_b \coloneq h_b$ for $b \neq a'$. Let $I$ be an arc of angles of length $\delta < 180^\circ$ such that for every $\theta$ outside of $I$, there exists no data-consistent \ac{scm} with outcome function $h^\theta$. Then within $I$, there is at most one $\theta$ for which a data-consistent \ac{scm} with outcome function $h^\theta$ exists.
\end{proposition}

Intuitively, if two rotations of the latent were both data consistent, then the rotation by their difference would be an element in $G_a$, so applying it repeatedly would produce further data-consistent rotations, and some of them would land among the rejected angles. The same argument applies in $\R^d$ to rotations within a fixed two-dimensional plane.

Suppose the \ac{ci} test rejects $h^\theta$ for all $\theta$ outside an arc $I$ of length $\delta < 180^\circ$ and does not reject inside $I$. By Proposition~\ref{pro:resolution}, at most one of the unrejected outcome functions is data consistent. All others are wrong maps that the test fails to detect, so the width $\delta$ reflects the power of the test and not the size of $G_a$. We illustrate this in our experiments in Section~\ref{sec:actg}, where we choose for $h$ the canonical outcome function of a selective class $\scm(\mathcal{C}, \eta)$ with $\eta=\normal(0, I_2)$. Then, $h$ satisfies the marginal transport condition by definition and so does every $h^\theta$ because $\eta$ is rotation invariant. Because every $h^\theta$ satisfies the marginal transport condition, $\delta$ informs us about the power of the test to detect whether an outcome function fails to satisfy the conditional transport conditions on the given data. We then refer to $\delta$ as the resolution of the test on the data and for the chosen $h$.
\section{Experiments}\label{sec:experiments}

In this section, we present two experiments. The first experiment uses synthetic data and illustrates how the identified set of counterfactuals varies with different levels of heterogeneity of $\nu_{a, v}$. The second experiment applies the \ac{ci} test (Theorem~\ref{thm:testability}) to real-world data and evaluates the resolution of the data along rotations (Proposition~\ref{pro:resolution}).

\subsection{Synthetic Data: Identified Sets and Power of the Test}\label{sec:toy-example}

\begin{figure}[ht]
    \centering
    \includegraphics[width=0.9\linewidth]{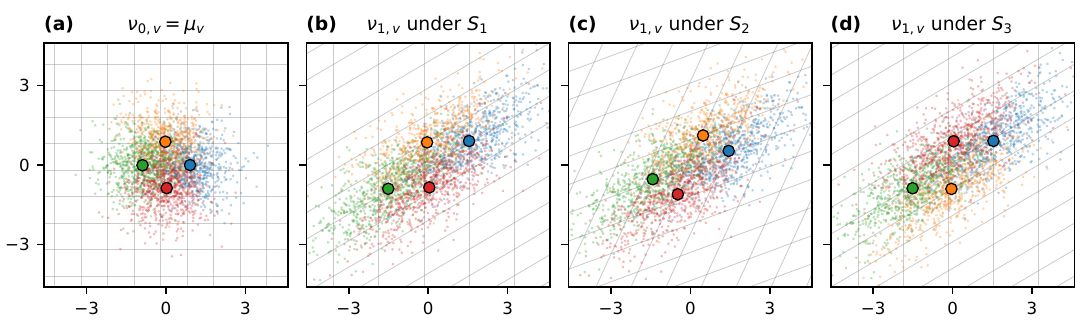}
    \caption{Observed distributions for the three different settings when $K=4$ coloured by $v$. (a) shows $\nu_{0,v}$ which is equal in all settings. (b--d) show $\nu_{1,v}$ for the three settings. Larger points denote the means of $\nu_{a,v}$.}
    \label{fig:toy-example-nu-a}
\end{figure}

We consider a two-dimensional outcome with a binary treatment $A \sim \bernoulli(1/2)$ and a latent $U \in \R^2$ with $\mu = \normal(0, I_2)$. The covariate $V$ takes $K$ values and partitions the latent space into $K$ angular sectors: $V$ indicates the sector of a noisy version of $U$, and a parameter $\rho \in [0, 1)$ controls how strongly the conditional distributions $\mu_v$ differ across sectors.
The outcome is generated as $Y = U$ for $A=0$ and $Y = S_i U$ for $A=1$, i.e. $f_0 = \id$ and $f_1 = S_i$, for one of three matrices $S_1, S_2, S_3$ that define three settings. The matrices induce the same interventional distributions $\nu_a$ and differ only in the conditional distributions $\nu_{1,v}$ (Figure~\ref{fig:toy-example-nu-a}). In Setting~1, $S_1$ is the \ac{kr} map in ordering $(1,2)$, so $f \in \triClass^{\state{A}}$. In Setting~2, $S_2$ is the Brenier map, so $f \in \otClass^{\state{A}}$. In Setting~3, $S_3 = S_1 R$ with the reflection $R = \diag(1, -1)$, so $f$ lies in neither class. The exact data generating process is given in Appendix~\ref{app:details-toy-example}.

\begin{figure}[ht]
    \centering
    \includegraphics[width=0.7\linewidth]{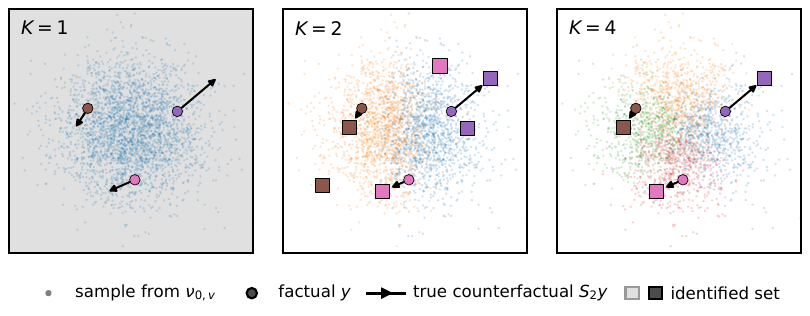}
    \caption{Identified sets in the synthetic example (Setting~2) for an increasing number of sectors $K$. The background shows the conditional distributions $\nu_{0,v}$, coloured by $v$. Three factual points $y$, and the identified set of the counterfactual of each $y$: the whole plane for $K=1$, two points for $K=2$, and the single point $S_2y$ for $K=4$.}
    \label{fig:toy-example-identified-set}
\end{figure}

\paragraph{Identified set.}
Figure~\ref{fig:toy-example-identified-set} shows how the identified set $T^f_{0 \to 1}(G_0 \cdot y) = S_i(G_0 \cdot y)$ shrinks as the number of angular sectors grows. According to Theorem~\ref{thm:symmetry-group}, $G_0$ depends only on the family $\myset{\nu_{0,v}}_v$, which is the same in all three settings. Therefore, while the figure shows Setting~2, the following remarks apply to all three settings.

For $K=1$, $V$ is constant and $G_0$ consists of all maps preserving $\nu_{0,1} = \normal(0, I_2)$. Any point can be moved to any other point by such a map, so $T^f_{0 \to 1}(G_0 \cdot y) = \R^2$ (Proposition~\ref{prop:no-conf-large-group}). For $K=2$, one can show (Appendix~\ref{app:details-toy-example}) that $G_0 = \myset{\id, R}$ with $R = \diag(1, -1)$, so the identified set has two points, $T^f_{0 \to 1}(G_0 \cdot y) = \myset{S_2 y, S_2 R y}$. Since $S_3 = S_1 R$, this also means that for $K = 2$, Settings~1 and~3 induce the same conditional distributions and cannot be distinguished by any test. For $K = 4$, no non-trivial map preserves all $\nu_{0,v}$ simultaneously, so $G_0 = \myset{\mathrm{id}}$ and counterfactuals are point identified.

\begin{figure}[ht]
    \centering
    \includegraphics[width=0.9\linewidth]{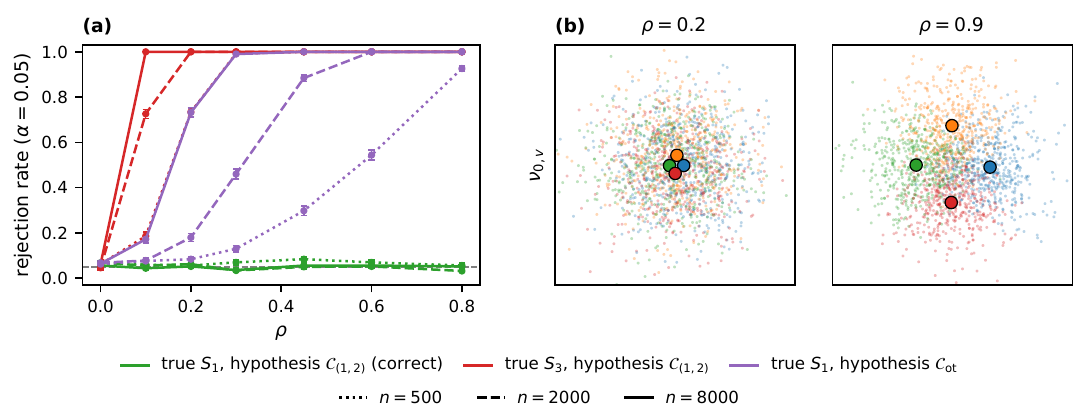}
    \caption{(a) Rejection rate of the \ac{ci} test against $\rho$ for three combinations of true outcome function and hypothesised model class and three sample sizes. (b) The conditional distributions $\nu_{0,v}$ for two values of $\rho$.}
    \label{fig:toy-example-power-rho}
\end{figure}

\paragraph{Power of the test.}
For $K=4$, $G_0$ is trivial, so any wrong outcome function is detectable. We evaluate the model classes $\scm(\triClass, \eta)$ with $\pi = (1,2)$ and $\scm(\otClass, \eta)$ with $\eta = \mu = \normal(0, I_2)$, i.e. with the correct reference distribution.
Figure~\ref{fig:toy-example-power-rho} shows how the rejection rate of the \ac{ci} test from Theorem~\ref{thm:testability} depends on how strongly the conditional distributions $\nu_{0,v}$ differ across $v$, which in our data generating process is controlled by $\rho$ (panel~(b)).
Since $V$ is discrete and $A$ binary, testing $(h^{\mathcal{C}}_A)^{-1}(Y) \indep A \mid V$ amounts to $K$ two-sample tests of
$\law{(h^{\mathcal{C}}_0)^{-1}(Y) \mid A=0, V=v} = \law{(h^{\mathcal{C}}_1)^{-1}(Y) \mid A=1, V=v}$,
one for each value $v$. We test for equal means with James' two-sample statistic in each stratum \citep{james_tests_1954} (details in Appendix~\ref{app:details-toy-example}).

Panel~(a) shows that larger $\rho$ increases power in much the same way as more samples do. When the hypothesised model class is correct (green: true $S_1$, hypothesis $\scm(\mathcal{C}_{(1,2)}, \eta)$), the rejection rate stays at the nominal level $\alpha = 0.05$. When the data is generated with $S_3$ but the hypothesised model class is $\mathcal{M}(\mathcal{C}_{(1,2)}, \eta)$ (red), then the class is rejected almost always for $\rho \ge 0.3$, even with $n = 500$. When the hypothesised model class is $\mathcal{M}(\otClass, \eta)$ but the true outcome function is $S_1$ (purple), rejection requires more data or larger $\rho$.
\subsection{Testing Structural Assumptions on ACTG~175}\label{sec:actg}

We evaluate the model classes $\mathcal{M}(\triClass, \eta)$ (in both orderings) and $\mathcal{M}(\otClass, \eta)$ with $\eta = \normal(0, I_2)$ on the ACTG~175 trial \citep{hammer_trial_1996}, which randomly assigned HIV-infected adults to different treatment arms. We compare zidovudine monotherapy (ZDV, $A=0$, $n_0 = 532$) with zidovudine plus didanosine (ZDV+ddI, $A=1$, $n_1 = 522$). The outcome $Y \in \R^2$ is the vector of CD4 and CD8 (log-transformed and standardised) cell counts 20 weeks after the start of treatment. The dataset also contains the CD4 and CD8 counts just before the start of the treatment, which are strongly associated with the outcome. We use them as $V$ under the assumption that they are associated with $Y$ only through the patient’s latent state $U$ (Assumption~\ref{ass:structural-assumptions}~\ref{ass:outcome-func-of-u-a}), which is plausible if $U$ represents the patient’s underlying immune status. We discretise these baseline CD4 and CD8 counts into terciles, so $V$ takes nine values. Since the treatment was randomised, $A \indep (U, V)$ and $\nu_a = \law{Y \mid A = a}$, so we have no confounding but heterogeneous conditional distributions $\nu_{0,v}$ (Appendix Figure~\ref{fig:actg-data}(b)).

\paragraph{Canonical counterfactual maps.}
For $\triClass$ we fit one autoregressive normalising flow per arm by maximum likelihood, in the orderings (CD4, CD8) and (CD8, CD4). For $\otClass$ we solve a discrete optimal transport problem between $n_a$ samples from $\eta$ and the observed outcomes of arm $a$ and extend the resulting discrete map to $\R^2$ by $k$-nearest-neighbour regression on the matched pairs ($k = 15$). The three counterfactual maps closely agree: all three predict that ZDV+ddI increases the CD4 count of almost every patient and leaves CD8 essentially unchanged (Appendix Figure~\ref{fig:actg-cf-maps}). The median distance between the \ac{kr} and Brenier predictions over the ZDV sample is $0.15$ standard deviations. The largest disagreement is at high CD4 and low CD8, where $\triClass{(\text{CD4, CD8})}$ predicts a decrease and $\otClass$ an increase in CD8, but there is little data in that region.

\begin{figure}[ht]
    \centering
    \includegraphics[width=0.9\linewidth]{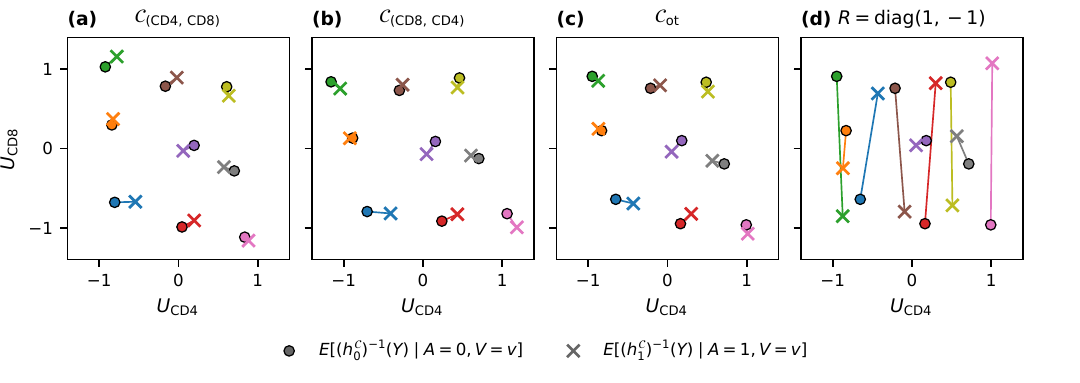}
    \caption{Pulled-back conditional means of the observed ZDV outcome (circles) and the observed ZDV+ddI outcome (crosses), coloured by $V$, for the three hypothesised classes (a--c) and the reflected Brenier map $T_R$ (d). A data-consistent map must pull back each treated mean onto its untreated counterpart (compare Figure~\ref{fig:testability-intuition} (b)). The reflection does not achieve that.}
    \label{fig:actg-pulled-back-strata}
\end{figure}

\paragraph{Testing data consistency.}
We apply the \ac{ci} test of Theorem~\ref{thm:testability} as in Section~\ref{sec:toy-example}, with one two-sample James' statistic per value of $V$. None of the three classes is rejected ($p = 0.92$, $0.85$, $0.90$ for $\triClass{(\text{CD4, CD8})}$, $\triClass{(\text{CD8, CD4})}$, $\otClass$). Figure~\ref{fig:actg-pulled-back-strata}(a--c) shows why: for each $v$, the pulled-back conditional ZDV mean $\E[(h^{\mathcal{C}}_0)^{-1}(Y) \mid A=0, V=v]$ lands on the pulled-back conditional ZDV+ddI mean $\E[(h^{\mathcal{C}}_1)^{-1}(Y) \mid A=1, V=v]$, which is implied by requirement~(ii) in Theorem~\ref{thm:testability} and exactly what we test for. Our test therefore only checks a necessary consequence of (ii) and has less power than a full distributional \ac{ci} test.

As discussed in Section~\ref{sec:testability}, non-rejection does not mean that the three maps are correct, only that each lies in $\myset{T^f_{0\to1} \circ M : M \in G_0}$. Since the three maps differ, either $G_0$ is non-trivial, or the test lacks the power to reject the wrong ones because they are too similar given the resolution of the test on the data. We can measure how different two maps have to be before the test distinguishes according to Proposition~\ref{pro:resolution}.

\paragraph{Power on real data.}
We construct new hypothesised outcome functions $h^R$ that agree with $\otClass$ on the marginals $\nu_0$ and $\nu_1$, but differ in the counterfactual map. For any orthogonal matrix $R$, we have that $R_\# \eta = \eta$, because $\eta = \normal(0, I_2)$ is invariant to rotations and reflections.
Similar to Proposition~\ref{pro:resolution}, we can define
$h^{R}_0 := h^{\otClass}_0, \ h^{R}_1 := h^{\otClass}_1 \circ R$ with counterfactual map $T_R \coloneq h^R_1 \circ (h^R_0)^{-1}$,
without changing the marginal transport $(h^R_0)_\# \eta = \nu_0$ and $(h^R_1)_\# \eta = \nu_1$. We can test if requirement~(ii) of Theorem~\ref{thm:testability} is satisfied using the same \ac{ci} test as before.
For example, for the reflection $R = \diag(1,-1)$, Figure~\ref{fig:actg-pulled-back-strata}(d) illustrates why the independence test rejects it ($p < 10^{-140}$): the pulled-back conditional means differ between the treatment arms.

We repeat this \ac{ci} test for rotation matrices $R_\theta$ (Appendix Figure~\ref{fig:actg-rotation-sweep}). The test does not reject for $|\theta| \le 10^\circ$, rejects at $|\theta| = 15^\circ$ for all three base classes except $\mathcal{C}_{(\mathrm{CD8},\mathrm{CD4})}$ at $\theta = 15^\circ$, and rejects with $p < 10^{-8}$ for $|\theta| \ge 30^\circ$. Proposition~\ref{pro:resolution} with $I \approx (-15^\circ, 15^\circ)$ then implies that at most one of the unrejected rotated maps is data consistent and the remaining unrejected rotations are wrong maps that the test fails to detect. The pulled-back data under the three maps differ from each other by a median rotation of $7^\circ$ to $14^\circ$, but the part of this rotation that differs between the two arms, which is all the test can detect, is less than $7^\circ$ for every pair of classes. This lies within $I$ and explains why all three maps pass the \ac{ci} test.

\section{Discussion}\label{sec:discussion}

Our results rely on the assumption that the covariate $V$ affects the outcome only through the latent $U$. This assumption enters the \ac{ci} test only jointly with the hypothesised outcome function, so a rejection may indicate a wrong model class or a violated assumption, and a non-rejection does not verify the assumption, since a direct effect of $V$ on $Y$ can leave the conditional transport condition satisfied for a map that is not the true counterfactual map. What matters beyond this is heterogeneity of the latent distribution across the values of $V$. It determines how much the data says about the counterfactual map. The covariate does not need to be a confounder for this, so randomised trials with informative baseline covariates are as useful as observational data.

A non-rejection can be due to a lack of power or to a genuine symmetry in $G_a$. Proposition~\ref{pro:resolution} lets us tell the two apart along rotations of the latent, but only in that direction and for the given outcome function. Symmetries that do not lie on a rotation path, such as reflections, have to be tested separately. In our experiments we only test for equal conditional means, which is a necessary consequence of the \ac{ci} condition, and more general kernel based \ac{ci} tests can be used without changing the theory. In the synthetic example we treat the canonical outcome functions as known, whereas in practice they are estimated, and we do not analyse how estimation error affects the test.

Future work may investigate how the elements of $G_a$ can be computed or bounded to allow for partial identification of counterfactuals. Another line of research may use an instrumental variable instead of the covariate $V$ similar to the control function approach of \citet{imbens_identification_2009}.


\bibliography{references.bib}

\appendix

\section{Proofs}

\subsection{Symmetry Group}

\subsubsection{Theorem~\ref{thm:symmetry-group}: Symmetry Group}

\begin{proof}
    Instead of for counterfactuals $a \to a'$, we write the proof for counterfactuals $a' \to a''$ without loss of generality.
    We first show that for every $M \in G_{a'}$ there exists an \ac{scm} $\scmalt$ with alternative outcome function $h$ that satisfies $h_{a''} \circ h_{a'}^{-1} = T^f_{a' \rightarrow a''} \circ M$ , and that $\scmalt$ is observationally equivalent to $\scm$ with outcome function $f$.

    We prove this claim by constructing an alternative outcome function $h: \state{A} \times \R^d \rightarrow \R^d$ as
    \[
    h_a =
    \begin{cases}
        f_a &\text{if } a\neq a', \\
        M^{-1} \circ f_{a'} &\text{if } a = a'.
    \end{cases}
    \]

    Because by assumption $a' \neq a''$, this outcome function then satisfies

    \begin{align*}
    h_{a''} \circ h_{a'}^{-1} &= f_{a''} \circ f_{a'}^{-1} \circ M \\
    &= T^f_{a' \rightarrow a''} \circ M
    \end{align*}

    by construction.
    We furthermore have that
    \begin{align*}
        {h_a}_\# \mu_v
            &= {f_a}_\# \mu_v
            = \nu_{a, v},
            && \text{for } a \neq a', \\
        {h_{a'}}_\# \mu_v
            &= M^{-1} \circ {f_{a'}}_\# \mu_v \\
            &= M^{-1}_\# \nu_{a', v} \\
            &= \nu_{a', v},
            && \text{because } M^{-1} \in G_{a'}
    \end{align*}

    which together implies ${h_a}_\# \mu_v = \nu_{a, v}$ for all $a \in \state{A}$ and $\law{V}(v)\almev\ v$ and therefore observational equivalence. The constructed \ac{scm} satisfies Assumption~\ref{ass:structural-assumptions}.

    We now show that for a given alternative outcome function $h$ and conditional exogenous distribution $\tilde \mu_v$ that define an observationally equivalent \ac{scm} $\scmalt$, there exists an $M \in G_{a'}$ with $h_{a''} \circ h_{a'}^{-1} = T^f_{a' \rightarrow a''} \circ M$.
    Let us define a candidate $M \coloneq T^{-1}_{a' \rightarrow a''} \circ h_{a''} \circ h_{a'}^{-1}$ (note that $M$ does indeed not vary with $v$). Then by definition $h_{a''} \circ h_{a'}^{-1} = T^f_{a' \rightarrow a''} \circ M$. It remains to show that $M \in G_{a'}$.
    Because of the observational equivalence of $\scmalt$ we have that
    \[
    {h_a}_\# \tilde \mu_v = \nu_{a, v} \qquad \text{for all } a \in \state{A}, \law{V}(v)\almev\ v.
    \]
    This implies
    \begin{align*}
        \bigl( h_{a''} \circ h_{a'}^{-1} \bigr)_\# \nu_{a', v}
        &= {h_{a''}}_\# \tilde \mu_v \\
        &= \nu_{a'', v}.
    \end{align*}
    But because of $h_{a''} \circ h_{a'}^{-1} = T^f_{a' \rightarrow a''} \circ M$, this implies that
    \[
    \bigl( T^f_{a' \rightarrow a''} \circ M \bigr)_\# \nu_{a', v} = \nu_{a'', v}.
    \]
    At the same time we have that ${T^f_{a' \rightarrow a''}}_\# \nu = \nu_{a'', v}$ if and only if $\nu = \nu_{a', v}$ because $T^f_{a' \rightarrow a''}$ is a measurable bijection with ${T^f_{a' \rightarrow a''}}_\# \nu_{a', v} = \nu_{a'', v}$.
    It therefore follows that $M_\# \nu_{a', v} = \nu_{a', v}$, which means exactly $M \in G_{a'}$.

    It remains to show that $(G_{a}, \circ)$ is a group. We show that it satisfies all properties of a group.
    \begin{enumerate}
        \item[(i)] \emph{Associativity.}
        Let $B,C,D \in G_a$. Then, for every $y \in \R^d$,
        \[
            \bigl(B \circ (C \circ D)\bigr)(y)
            = B(C(D(y)))
            = \bigl((B \circ C) \circ D\bigr)(y).
        \]
        We therefore have
        \[
            B \circ (C \circ D) = (B \circ C) \circ D.
        \]

        \item[(ii)] \emph{Identity element.}
        Let
        \[
            \id \colon \R^d \to \R^d,
            \qquad y \mapsto y,
        \]
        denote the identity map. Clearly,
        \[
            \id \in \Diffeo^1(\R^d),
        \]
        and
        \[
            \id_\# \nu_{a, v} = \nu_{a, v} \qquad \text{for all } v.
        \]
        Thus, $\id \in G_a$. Moreover, for every $D \in G_a$ we have
        \[
            D \circ \id = D = \id \circ D.
        \]
        Therefore, $\id$ is the identity element of $G_a$.

        \item[(iii)] \emph{Inverse elements.}
        Let $D \in G_a$. Since
        \[
            D \in \Diffeo^1(\R^d),
        \]
        its inverse satisfies
        \[
            D^{-1} \in \Diffeo^1(\R^d).
        \]
        Furthermore, since
        \[
            D_\# \nu_{a, v} = \nu_{a, v},
        \]
        applying $(D^{-1})_\#$ to both sides gives
        \[
            (D^{-1})_\# \nu_{a, v}
            = (D^{-1})_\#(D_\# \nu_{a, v})
            = (D^{-1} \circ D)_\# \nu_{a, v}
            = \nu_{a, v}.
        \]
        Hence, $D^{-1} \in G_a$.

        \item[(iv)] \emph{Closure under composition.}
        Let $C,D \in G_a$. Since $C$ and $D$ are $C^1$-diffeomorphisms,
        \[
            C \circ D \in \Diffeo^1(\R^d).
        \]
        Moreover,
        \[
            (C \circ D)_\# \nu_{a, v}
            = C_\#\bigl(D_\# \nu_{a, v}\bigr)
            = C_\# \nu_{a, v}
            = \nu_{a, v}.
        \]
        Therefore,
        \[
            C \circ D \in G_a.
        \]
    \end{enumerate}
\end{proof}

\subsubsection{Point-identification under trivial symmetry}
We now prove the claim that counterfactuals are point identified if and only if $G_a$ is trivial.
\begin{proof}
    ($\implies$) Let counterfactuals be point identified and let us assume there exists $M \in G_a$ with $M \neq \id$. Then there exists a $y \in \R^d$ such that $M(y) \neq y$. Consequently we have for $y' \coloneq T^f_{a \rightarrow a'}(y)$ that $T^f_{a \rightarrow a'}(M(y)) \neq y'$ because of the bijectivity of $T^f_{a \rightarrow a'}$. Therefore, counterfactuals are not point identified which is a contradiction.

    ($\impliedby$) Let $G_a = \myset{\id}$. Fix $a$ and $a'$. Then according to Theorem~\ref{thm:symmetry-group}, all counterfactual maps $a \rightarrow a'$ achievable by data consistent \acp{scm} are exactly $\myset{T^f_{a \rightarrow a'} \circ \id} = \myset{T^f_{a \rightarrow a'}}$. Because the set contains only one element, counterfactuals are point identified.
\end{proof}

\subsubsection{Proposition~\ref{pro:nasr}: Nasr-Esfahany}

\begin{proof}
    Let $a$ be fixed, $M \in G_a$, and $S \coloneq f_a^{-1} \circ M \circ f_a$.

    Then
    \begin{align*}
        S_\# \mu_v &= (f_a^{-1} \circ M \circ f_a)_\# \mu_v \\
        &= (f_a^{-1} \circ M)_\# \nu_{a, v} \\
        &= {f_a^{-1}}_\# \nu_{a, v} \\
        &= \mu_v.
    \end{align*}

    We know that $S \in \Diffeo^1(\R^d)$ and $S_\# \mu_v = \mu_v$ for all $v \in \state{V}$. Note that if we can show that $S = \id$, then it immediately follows that $M = \id$ and $G_a$ is trivial.
    
    Let $p_v$ be the pdf of $U \mid V=v$. According to the change of variables formula we then have
    \begin{align}\label{eq:change-of-variables}
    p_v(S(u)) \cdot \left| \det J_{S(u)} \right| = p_v(u) \qquad \text{for all } u, v.
    \end{align}
    This equality implies that if $p_v(u) \neq 0$, then $p_v(S(u)) \neq 0$. If for any $j \in \myset{0, \dots, d}$ and any $u$ we had $p_{v_j}(u) = 0$, then $p_{v_j}$ would have a local minimum in $u$ and we would therefore have $D p_{v_j}(u) = 0$. Then, the $j+1$-th row would consist only of zero entries so the non-zero determinant assumption would be violated. We therefore conclude that $p_{v_j}(u) > 0$ for all $u, j$.
    If we then divide Equation~\ref{eq:change-of-variables} for $v_i$ by itself for $v_0$, we get
    \begin{equation}\label{eq:prob-ratio-preserve}
        \frac{p_{v_i}(S(u))}{p_{v_0}(S(u))} = \frac{p_{v_i}(u)}{p_{v_0}(u)}.
    \end{equation}

    If we then define 
    \[
    R: \R^d \rightarrow \R^d, u \mapsto R(u) \coloneq \begin{pmatrix}
        p_{v_1}(u) / p_{v_0}(u) \\
        \vdots \\
        p_{v_d}(u) / p_{v_0}(u)
    \end{pmatrix},
    \]
    then Equation~\ref{eq:prob-ratio-preserve} implies
    \begin{equation}\label{eq:r-fix-s}
        R(S(u)) = R(u) \qquad \text{for all } u \in \R^d.
    \end{equation}
    We observe that if $R$ is injective, then Equation~\ref{eq:r-fix-s} already implies $S = \id$.

    Note that if $S_\# \mu_v = \mu_v$ for all $v$, then also $S^k \coloneq S \circ S \dots \circ S$ $k$-times preserves the law. We therefore also have
    \begin{equation}\label{eq:power-k-r-fix-s}
        R(S^k(u)) = R(u) \qquad \text{for all } u \in \R^d.
    \end{equation}
    
    We will now show that for each $u_0 \in \R^d$, $R$ is locally injective, i.e. there exists a neighbourhood of $u_0$ restricted to which $R$ is injective.

    According to our assumption we have
    \begin{equation}\label{eq:non-zero-det}
        \det \begin{pmatrix}
            p_{v_0}(u) & D p_{v_0}(u) \\
            \vdots & \vdots \\
            p_{v_d}(u) & D p_{v_d}(u)
        \end{pmatrix}
        \neq 0 \qquad \text{for all } u.
    \end{equation}
    At the same time we can use basic rules for calculating determinants to show that $\det D R(u) \neq 0$ for all $u \in \R^d$. For each row $i \in \{2, \dots, d+1\}$ we first subtract $\frac{p_{v_i}(u)}{p_{v_0}(u)}$ times the first row to get
    \begin{align}
        \det \begin{pmatrix}
            p_{v_0}(u) & D p_{v_0}(u) \\
            \vdots & \vdots \\
            p_{v_d}(u) & D p_{v_d}(u)
        \end{pmatrix}
        &= 
        \det \begin{pmatrix}
            p_{v_0}(u) & D p_{v_0}(u) \\
            0 & D p_{v_1}(u) - \frac{p_{v_1}(u)}{p_{v_0}(u)} D p_{v_0}(u) \\
            \vdots & \vdots \\
            0 & D p_{v_d}(u) - \frac{p_{v_d}(u)}{p_{v_0}(u)} D p_{v_0}(u)
        \end{pmatrix}.
    \end{align}
    At the same time we have
    \begin{align*}
        D \frac{p_{v_i}}{p_{v_0}}(u)
        &=
        \frac{D p_{v_i} (u)}{p_{v_0}(u)} - p_{v_i}(u) \cdot D p_{v_0}(u) \cdot \frac{1}{(p_{v_0}(u))^2} \\
        &=
        \frac{1}{p_{v_0}(u)} \left( D p_{v_i}(u) - \frac{p_{v_i}(u)}{p_{v_0}(u)} D p_{v_0}(u) \right)\\
        \iff p_{v_0}(u) \cdot D \frac{p_{v_i}}{p_{v_0}}(u) &= \left( D p_{v_i}(u) - \frac{p_{v_i}(u)}{p_{v_0}(u)} D p_{v_0}(u) \right)
    \end{align*}
    which implies
    \begin{equation}
        \det \begin{pmatrix}
            p_{v_0}(u) & D p_{v_0}(u) \\
            \vdots & \vdots \\
            p_{v_d}(u) & D p_{v_d}(u)
        \end{pmatrix}
        = 
        \det \begin{pmatrix}
            p_{v_0}(u) & D p_{v_0}(u) \\
            0 & p_{v_0}(u) \cdot D \frac{p_{v_1}}{p_{v_0}}(u) \\
            \vdots & \vdots \\
            0 & p_{v_0}(u) \cdot D \frac{p_{v_d}}{p_{v_0}}(u)
        \end{pmatrix}.
    \end{equation}

    Using the Laplace expansion for the first column gives us
    \begin{align}
        (0 \neq)
        \det \begin{pmatrix}
            p_{v_0}(u) & D p_{v_0}(u) \\
            \vdots & \vdots \\
            p_{v_d}(u) & D p_{v_d}(u)
        \end{pmatrix}
        &= 
        p_{v_0}(u) \cdot
        \det \begin{pmatrix}
            p_{v_0}(u) \cdot D \frac{p_{v_1}}{p_{v_0}}(u) \\
            \vdots \\
            p_{v_0}(u) \cdot D \frac{p_{v_d}}{p_{v_0}}(u)
        \end{pmatrix} \\
        &=
        \left(p_{v_0}(u)\right)^{d+1} \cdot
        \det \begin{pmatrix}
            D \frac{p_{v_1}}{p_{v_0}}(u) \\
            \vdots \\
            D \frac{p_{v_d}}{p_{v_0}}(u)
        \end{pmatrix} \\
        &=
        \left(p_{v_0}(u)\right)^{d+1} \cdot \det D R(u).
    \end{align}
    Consequently, we can conclude that $\det D R(u) \neq 0$ and the inverse function theorem therefore implies that $R$ is locally injective everywhere.

    We now show the following lemma.
    \begin{lemma}\label{lem:if-fixed-point-then-id}
    if there is a fixed point $u_0$ of $S^k$, i.e. $S^k(u_0) = u_0$, then every point in $\R^d$ is a fixed point and consequently $S^k = \id$.
    \end{lemma}
    
    Because $S^k$ is continuous, for every $\varepsilon > 0$ we can find a neighbourhood $W$ of $u_0$ such that
    \[
    \norm{S^k(u) - S^k(u_0)} = \norm{S^k(u) - u_0} < \varepsilon \qquad \text{whenever } u \in W.
    \]
    That means that we can find a neighbourhood $W$ of $u_0$ that is small enough such that it is a subset of a neighbourhood $N$ of $u_0$ for which $R$ restricted to $N$ is injective.
    According to Equation~\ref{eq:power-k-r-fix-s} we then have
    \[
    S^k(u) = u \qquad \text{for } u \in W.
    \]
    In other words, every point in $W$ is also a fixed point. Because the initial fixed point was arbitrary, that means that for any fixed point $u_0 \in \R^d$ we can find a neighbourhood that only contains fixed points. This implies that the fixed point set is open.

    At the same time we have that the fixed point set is closed which can be seen as follows. Let $\nu = S^k - \id$, i.e. $\nu$ is continuous with $\nu(u) = S^k(u) - u$. Then the fixed point set is exactly
    \[
    \myset{u : S^k(u) = u} = \myset{u: \nu(u) = 0} = \nu^{-1}(\myset{0}).
    \]
    Because $\nu$ is continuous and $\myset{0}$ is closed we conclude that the fixed point set must be closed.

    The only sets in $\R^d$ that are both open and closed are the empty set $\emptyset$ and $\R^d$ itself, and because by assumption $u_0$ is in the fixed point set, we can conclude that the fixed point set is equal to $\R^d$ and consequently $S^k = \id$.

    We will now show that $S$ has finite order, i.e. there exists a $k \in \N$ such that $S^k = \id$.
    Let $N$ be the neighbourhood of a point $u_0$ where $R$ is injective. Because of $p_{v_0}(u) > 0$ we then have $\mu_{v_0}(N) > 0$. There are now two possibilities: Either 1. the sets $\dots S^{-1}(N), N, S(N), S^2(N), \dots$ are pairwise disjoint or 2. they are not pairwise disjoint.

    In the first case we have that $\dots S^{-1}(N), N, S(N), S^2(N), \dots$ are pairwise disjoint.
    Because $S$ preserves the distribution of $\mu_{v_0}$ we have that $\mu_{v_0}(S^n(N)) = \mu_{v_0}(N) > 0$ for all $n \in \N$. Because the aforementioned sets are disjoint, we have that
    \[
    1 \geq \mu_{v_0} \left(\bigcup_{n=0}^\infty S^n(N) \right) = \sum_{n=0}^\infty \mu_{v_0} \left( S^n(N) \right) = \sum_{n=0}^\infty \mu_{v_0} \left( N \right) = \mu_{v_0} \left( N\right) \cdot \sum_{n=0}^\infty 1 = \infty
    \]

    which is a contradiction. Hence we can conclude that we are in case 2 where the sets are not pairwise disjoint.

    Then there exist $n, m$ with $n \neq m$ such that
    \[
    S^m(N) \cap S^n(N) \neq \emptyset \iff N \cap S^{n-m}(N) \neq \emptyset.
    \]
    So there exists $x, y \in N$ such that
    \[
    y = S^{n-m}(x).
    \]
    By Equation~\ref{eq:power-k-r-fix-s} we then have
    \[
    R(y) = R(x)
    \]
    and because $R$ is injective on $N$ this implies $x = y$. Therefore, $S^{n-m}$ has a fixed point and according to Lemma~\ref{lem:if-fixed-point-then-id} it then follows for $k = n - m$ that $S^k = \id$.

    It now remains to show that the order of $S$ is exactly $1$, to rule out finite-order symmetries (e.g. for $f(x) = -x$ we have $f \circ f = \id$ but $f \neq \id$).
    We show this by contradiction and assume that the order of $S$ is $n > 1$ (i.e. $S \neq \id)$. There then exists a prime $p$ that divides $n$ and we define $T \coloneq S^{n/p}$. Then $T$ has order $p$ because $T^p = S^n = \id$ and $T^j = S^{j \frac{n}{p}} \neq \id$ for $j < p$. Then $\myset{\id, T, \dots, T^{p-1}}$ is a p-group that acts on $\R^d$ and according to Smith fixed point theory there then exists a fixed point $u^\star \in \R^d$ with $T(u^\star) = u^\star$ \citep{smith_theorem_1934}. But according to Lemma~\ref{lem:if-fixed-point-then-id} it then follows that $T = S^{\frac{n}{p}} = \id$ which contradicts that $T$ has order $p > 1$. We therefore have $n=1$ and $S = \id$.

    $S = \id$ immediately implies $M = \id$ and because $M$ was chosen arbitrarily in $G_a$, this implies that $G_a = \myset{\id}$ is trivial.
\end{proof}

\subsubsection{Proposition~\ref{prop:no-conf-large-group}: No heterogeneity}

\begin{proof}
Under the hypothesis, $G_a = \{M \in \Diffeo^1(\R^d) : M_\# \nu_a = \nu_a\}$.
Since $\nu_a$ has a smooth positive density and $d \ge 2$, the diffeomorphisms of $\R^d$
preserving $\nu_a$ act transitively on $\R^d$ \citep{boothby_transitivity_1969}: for every $y' \in \R^d$
there is $M \in G_a$ with $M(y) = y'$. Hence $G_a \cdot y = \R^d$, and since
$T^f_{a \to a'}$ is a bijection of $\R^d$, $T^f_{a \to a'}(G_a \cdot y) = \R^d$.
\end{proof}
\subsection{Testability}

We start by recalling two well established results.

\begin{lemma}[\citet{bogachev_triangular_2005}]\label{lem:kr}
   Let $\eta, \nu$ be absolutely continuous on $\R^d$. There is a $\eta \almev$\ unique measurable triangular monotone map $g$ with $g_\# \eta = \nu$. In particular, at most one element of $\triClass$ transports $\eta$ to $\nu$, up to $\eta \almev$\ equality. Moreover, $\triClass$ is closed under composition and inversion.
\end{lemma}

\begin{lemma}[\citet{mccann_existence_1995}]\label{lem:brenier}
   Let $\eta$ be absolutely continuous on $\R^d$ and $\nu$ any distribution on $\R^d$. There is a $\eta \almev$\ unique gradient of a convex function $g$ with $g_\# \eta = \nu$. In particular, at most one element of $\otClass$ transports $\eta$ to $\nu$, up to $\eta \almev$\ equality.
\end{lemma}

\subsubsection{\texorpdfstring{$\triClass$}{triangular-monotone-class} and \texorpdfstring{$\otClass$}{OT-class} are selective, and the KR counterfactual map does not depend on the reference distribution}\label{app:tri-ot-selective}

\begin{proof}
   Let $\eta$ be absolutely continuous. Selectivity of $\scm(\triClass, \eta)$ and $\scm(\otClass, \eta)$ is the uniqueness statement in Lemmas~\ref{lem:kr} and~\ref{lem:brenier}, applied with $\nu = \nu_a$ for each $a \in \state{A}$.

   For the second claim, let $\eta$ and $\tilde\eta$ be absolutely continuous and let $f$ and $h$ be the canonical outcome functions of $\scm(\triClass, \eta)$ and $\scm(\triClass, \tilde\eta)$, assumed to exist. Then ${f_a}_\# \eta = \nu_a = {h_a}_\# \tilde\eta$ for all $a$. The map $\rho_a \coloneq h_a^{-1} \circ f_a$ lies in $\triClass$ by closure under composition and inversion, and ${\rho_a}_\# \eta = (h_a^{-1})_\# \nu_a = \tilde\eta$ for every $a$. By Lemma~\ref{lem:kr}, $\rho_a = \rho_{a'}$ $\eta \almev$\ for all $a, a'$, hence
   \[
   T^f_{a \to a'} = f_{a'} \circ f_a^{-1} = h_{a'} \circ \rho_{a'} \circ \rho_a^{-1} \circ h_a^{-1} = h_{a'} \circ h_a^{-1} = T^h_{a \to a'} \qquad \nu_a \almev
   \]
   That the analogous statement fails for $\otClass$ is shown by the counterexample in Appendix~\ref{app:ot-counterexample}.
\end{proof}

\subsubsection{Remark~\ref{rem:ot-needs-fixed-reference}: the Brenier counterfactual map depends on the reference distribution}
\label{app:ot-counterexample}

Let $d = 2$, $\state{A} = \{0,1\}$, $A \sim \mathrm{Bernoulli}(1/2)$, and let $V$ be constant, so that Assumption~\ref{ass:structural-assumptions}~(ii) and (iv) hold trivially. We use that the Brenier map between centred Gaussians $\mathcal{N}(0, \Sigma)$ and $\mathcal{N}(0, \Sigma')$ is the linear map $y \mapsto By$ with the unique symmetric positive definite $B$ satisfying $B\Sigma B = \Sigma'$, namely $B = \Sigma^{-1/2}(\Sigma^{1/2}\Sigma'\Sigma^{1/2})^{1/2}\Sigma^{-1/2}$. Every such map is the gradient of the convex function $y \mapsto \tfrac12 y^\top B y$ and a $C^1$-diffeomorphism, hence in $\otClass$.

Define
\[
  D \coloneq \begin{pmatrix} 2 & 0 \\ 0 & 1 \end{pmatrix}, \qquad
  C \coloneq \begin{pmatrix} 2 & 1 \\ 1 & 1 \end{pmatrix}, \qquad
  \Sigma_1 \coloneq D^{-1} C^2 D^{-1} = \begin{pmatrix} 5/4 & 3/2 \\ 3/2 & 2 \end{pmatrix},
\]
and let the observed data be $\nu_0 = \mathcal{N}(0, I_2)$ and $\nu_1 = \mathcal{N}(0, \Sigma_1)$.

\emph{First \ac{scm}.} Let $\eta \coloneq \mathcal{N}(0, I_2)$ and $f_0 \coloneq \mathrm{id}$, $f_1 \coloneq \Sigma_1^{1/2}$. Then $f \in \otClass^{\state{A}}$ and $(f_a)_\#\eta = \nu_a$ for $a = 0, 1$, so the \ac{scm} with latent distribution $\eta$ and outcome function $f$ lies in $\scm(\otClass, \eta)$ and is data consistent. Its counterfactual map is
\[
  T^f_{0 \to 1} = f_1 \circ f_0^{-1} = \Sigma_1^{1/2},
\]
a symmetric matrix.

\emph{Second \ac{scm}.} Let $\tilde\eta \coloneq \mathcal{N}(0, D^2) \neq \eta$ and $h_0 \coloneq D^{-1}$, $h_1 \coloneq D^{-1} C D^{-1}$. Both matrices are symmetric positive definite ($C$ has trace $3$ and determinant $1$), and
\[
  h_0 D^2 h_0 = I_2, \qquad h_1 D^2 h_1 = D^{-1} C^2 D^{-1} = \Sigma_1,
\]
so $h_a$ is the Brenier map from $\tilde\eta$ to $\nu_a$, $h \in \otClass^{\state{A}}$, and the \ac{scm} with latent distribution $\tilde\eta$ and outcome function $h$ lies in $\scm(\otClass, \tilde\eta)$ and is data consistent. Its counterfactual map is
\[
  T^h_{0 \to 1} = h_1 \circ h_0^{-1} = D^{-1} C D^{-1} D = D^{-1} C
  = \begin{pmatrix} 1 & 1/2 \\ 1 & 1 \end{pmatrix},
\]
which is not symmetric. Hence $T^h_{0\to 1} \neq T^f_{0 \to 1}$, although both \acp{scm} have Brenier outcome functions and induce the same observed distribution.

The mechanism is the one described in Remark~\ref{rem:ot-needs-fixed-reference}. The Brenier map $\rho \coloneq D$ transports $\eta$ to $\tilde\eta$, and $h_1 \circ \rho = D^{-1} C$ transports $\eta$ to $\nu_1$, but it is not symmetric and hence not in $\otClass$. The Brenier map from $\eta$ to $\nu_1$ is $\Sigma_1^{1/2}$ instead. In one dimension this cannot happen, since $\otClass$ then coincides with the increasing maps and is closed under composition.

\subsubsection{Theorem~\ref{thm:testability}: Testability}\label{app:testability}

\begin{proof}
    Throughout, the conditional distributions $\nu_{a,v}$ are defined for every $a \in \state{A}$ and $P_V \almev\ v$, because $\state{A}$ is finite and $P(A = a \mid V = v) > 0$ for $P_V \almev\ v$ by overlap (Assumption~\ref{ass:structural-assumptions}~\ref{ass:overlap}). For the same reason, a statement that holds for $P_{A,V} \almev\ (a,v)$ holds for every $a \in \state{A}$ and $P_V \almev\ v$.

    \textit{(i) $\implies$ (ii).}
    Let $\scmalt$ be a data-consistent \ac{scm} with outcome function $h$ and latent $\tilde U$, so that $\tilde Y = h_A(\tilde U)$ and, by Assumption~\ref{ass:structural-assumptions}~\ref{ass:u-indep-a-given-v}, $\tilde U \indep A \mid V$. Since $\scmalt$ induces the observed distribution of $(V, A, Y)$, the distribution of $h_A^{-1}(Y)$ under the observed distribution equals the distribution of $h_A^{-1}(\tilde Y) = \tilde U$ under $\scmalt$. Hence $h_A^{-1}(Y) \indep A \mid V$.

    \textit{(ii) $\implies$ (i).}
    The conditional distribution of $h_A^{-1}(Y)$ given $A = a$ and $V = v$ is $(h_a^{-1})_\# \nu_{a,v}$. By (ii) it does not depend on $a$, so we may define $\tilde \eta_v \coloneq (h_a^{-1})_\# \nu_{a,v}$ for $P_V \almev\ v$, independently of the choice of $a$. Let $\scmalt$ be an \ac{scm} with $\tilde V = V$, $\tilde A = A$, $\law{\tilde U \mid A = a, V = v} = \tilde\eta_v$ for all $a$, and $\tilde Y = h_{\tilde A}(\tilde U)$, where $\law{A, V}$ is the observed distribution. By construction, $\tilde U \indep A \mid V$, $h_a \in \Diffeo^1(\R^d)$ for all $a$, and overlap holds, so $\scmalt$ satisfies Assumption~\ref{ass:structural-assumptions}. Moreover,
    \[
    \law{\tilde Y \mid \tilde A = a, \tilde V = v} = {h_a}_\# \tilde\eta_v = {h_a}_\# (h_a^{-1})_\# \nu_{a,v} = \nu_{a,v},
    \]
    so $\scmalt$ induces the observed distribution of $(V, A, Y)$ and is data consistent.

    \textit{(i) $\implies$ (iii).}
    A data-consistent \ac{scm} with outcome function $h$ has counterfactual maps $T^h_{a \to a'}$, and Theorem~\ref{thm:symmetry-group} states that the counterfactual maps of data-consistent \acp{scm} lie in $\myset{T^f_{a \to a'} \circ M : M \in G_a}$.

    \textit{(iii) $\implies$ (ii).}
    Fix $a \in \state{A}$. For every $a'$ there is $M_{a'} \in G_a$ with $T^h_{a \to a'} = T^f_{a \to a'} \circ M_{a'}$. Since $M_{a'}$ preserves $\nu_{a,v}$ and $T^f_{a \to a'}$ transports $\nu_{a,v}$ to $\nu_{a',v}$ (both for $P_V \almev\ v$), we have for $P_V \almev\ v$
    \[
    (h_{a'}^{-1})_\# \nu_{a',v}
    = (h_{a'}^{-1})_\# (T^f_{a \to a'})_\# (M_{a'})_\# \nu_{a,v}
    = \big(h_{a'}^{-1} \circ T^h_{a \to a'}\big)_\# \nu_{a,v}
    = (h_a^{-1})_\# \nu_{a,v}.
    \]
    The left-hand side is the conditional distribution of $h_A^{-1}(Y)$ given $A = a'$ and $V = v$, and the right-hand side the one given $A = a$ and $V = v$. As $a'$ was arbitrary, this conditional distribution does not depend on the treatment, which is (ii). Note that this direction uses (iii) only for a single fixed $a \in \state{A}$ and all $a' \in \state{A}$.
\end{proof}

\subsubsection{Consequences for selective classes}\label{app:selective-consequences}

We prove the two claims made in Section~\ref{sec:testability} about selective classes.

\begin{proof}[Proof that the canonical outcome function is the only candidate]
    Let $\scm(\mathcal{C}, \eta)$ be selective and let $\scmalt \in \scm(\mathcal{C}, \eta)$ be data consistent with outcome function $h$ and latent $\tilde U$. Then $\law{\tilde U} = \eta$ by definition of the model class, and $\nu_a = \law{Y \mid \doop(A = a)} = {h_a}_\# \eta$ for every $a$ by the same argument as in Section~\ref{sec:background-and-problem-setup}, applied to $\scmalt$. Since $h_a \in \mathcal{C}$ and the class is selective, $h_a = h^{\mathcal{C}}_a$ $\eta \almev$\ for all $a$.
\end{proof}

\begin{proof}[Proof that without heterogeneity every selective class contains a data-consistent \ac{scm}]
    Let $U \indep V$, so that $\mu_v = \mu$ and hence $\nu_{a,v} = (f_a)_\# \mu_v = \nu_a$ for every $a$ and $P_V \almev\ v$. Let $\scm(\mathcal{C}, \eta)$ be selective with canonical outcome function $h^{\mathcal{C}}$. Then for every $a$ and $P_V \almev\ v$,
    \[
    \big((h^{\mathcal{C}}_a)^{-1}\big)_\# \nu_{a,v} = \big((h^{\mathcal{C}}_a)^{-1}\big)_\# \nu_a = \eta,
    \]
    so the conditional distribution of $(h^{\mathcal{C}}_A)^{-1}(Y)$ given $A = a$ and $V = v$ equals $\eta$ for all $a$, which is condition~(ii) of Theorem~\ref{thm:testability} for $h^{\mathcal{C}}$. By Theorem~\ref{thm:testability} there is a data-consistent \ac{scm} with outcome function $h^{\mathcal{C}}$, and the one constructed in the proof of (ii) $\implies$ (i) has $\law{\tilde U \mid V = v} = \eta$ for $P_V \almev\ v$, hence $\law{\tilde U} = \eta$. It therefore lies in $\scm(\mathcal{C}, \eta)$.
\end{proof}

\subsubsection{Proposition~\ref{pro:resolution}: Resolution along rotations}\label{app:resolution}

\begin{proof}
    Let $N_\theta \coloneq h_a \circ R_\theta \circ h_a^{-1}$. Then
    \[
    T^{h^\theta}_{a \to a'} = h_{a'} \circ R_\theta \circ h_a^{-1} = T^h_{a \to a'} \circ N_\theta,
    \]
    and $N_\theta \circ N_{\theta'} = N_{\theta + \theta'}$ for all $\theta, \theta'$, because $R_\theta \circ R_{\theta'} = R_{\theta + \theta'}$.

    Suppose that for some $\theta_1 \neq \theta_2$ in $I$ there are data-consistent \acp{scm} with outcome functions $h^{\theta_1}$ and $h^{\theta_2}$. By Theorem~\ref{thm:testability}, (i) $\implies$ (iii), there are $M_1, M_2 \in G_a$ with
    \[
    T^h_{a \to a'} \circ N_{\theta_1} = T^f_{a \to a'} \circ M_1
    \qquad\text{and}\qquad
    T^h_{a \to a'} \circ N_{\theta_2} = T^f_{a \to a'} \circ M_2 .
    \]
    Writing $\alpha \coloneq \theta_2 - \theta_1$, these give
    \[
    N_\alpha = N_{\theta_1}^{-1} \circ N_{\theta_2} = M_1^{-1} \circ M_2 \in G_a ,
    \]
    since $G_a$ is a group (Theorem~\ref{thm:symmetry-group}), and hence $N_{k\alpha} = N_\alpha^{\,k} \in G_a$ for every $k \in \mathbb{Z}$. Therefore
    \[
    T^{h^{\theta_1 + k\alpha}}_{a \to a'}
    = T^h_{a \to a'} \circ N_{\theta_1} \circ N_{k\alpha}
    = T^f_{a \to a'} \circ (M_1 \circ N_{k\alpha})
    \qquad\text{with } M_1 \circ N_{k\alpha} \in G_a ,
    \]
    For every other treatment $b \neq a'$ we have $h^{\theta_1 + k\alpha}_b = h^{\theta_1}_b$, so $T^{h^{\theta_1 + k\alpha}}_{a \to b} = T^{h^{\theta_1}}_{a \to b} \in \myset{T^f_{a \to b} \circ M : M \in G_a}$ by data consistency of $h^{\theta_1}$ and Theorem~\ref{thm:testability}, (i) $\implies$ (iii). Hence condition~(iii) of Theorem~\ref{thm:testability} holds for the fixed $a$ and all targets, which suffices for (iii) $\implies$ (ii) $\implies$ (i) by the note at the end of its proof. A data-consistent \ac{scm} with outcome function $h^{\theta_1 + k\alpha}$ therefore exists for every $k \in \mathbb{Z}$.

    Measured along the circle, $0 < |\alpha| \le \delta < 180^\circ$, while the complement of $I$ is an arc of length $360^\circ - \delta > 180^\circ > |\alpha|$. Consecutive angles $\theta_1 + k\alpha$ differ by $|\alpha|$, which is smaller than the length of that arc, so some $\theta_1 + k\alpha$ lies outside $I$. This contradicts the assumption that no data-consistent \ac{scm} with outcome function $h^\theta$ exists for $\theta \notin I$. Hence at most one $\theta \in I$ admits a data-consistent \ac{scm} with outcome function $h^\theta$.
\end{proof}

\section{Experiments}

\anonalt{The code and data for all experiments will be linked in de-anonymized version.}
        {The code and data for all experiments can be found under \url{https://github.com/marbr987/symmetries_of_counterfactuals}.}

\subsection{Synthetic Example}\label{app:details-toy-example}

\paragraph{Data generating process.}
We fix $K \in \N$ and $\rho \in [0,1)$ and partition $\R^2$ into $K$ angular sectors of equal angle $2\pi/K$. The data is generated as
\begin{align*}
  W &\sim \normal(0, I_2), & V &= \text{index in } \myset{1,\dots,K} \text{ of the sector containing } W,\\
  \varepsilon &\sim \normal(0, I_2), & U &= \rho W + \sqrt{1-\rho^2}\,\varepsilon,\\
  A &\sim \bernoulli(1/2), & Y &= f(A,U) := S_i^A U,
\end{align*}

with $W$, $\varepsilon$, and $A$ mutually independent. Hence $\mu = \normal(0, I_2)$ and $A \indep (U, V)$, so there is no confounding in this example. The heterogeneity of $\nu_{a, v}$ in $v$ arises solely because $U$ depends on $V$ through $W$. The conditional latent distributions $\mu_v$ are noisy versions of the sectors of $W$ (Figure~\ref{fig:toy-example-nu-a}(a)), and $\rho$ controls how strongly they differ across $v$. Since $f_0 = \mathrm{id}$ and $f_1 = S_i$, we have $\nu_{0,v} = \mu_v$, $\nu_{1,v} = (S_i)_\# \mu_v$, and $T^f_{0\to 1} = S_i$. The index $i \in \myset{1,2,3}$ selects one of three outcome functions,
\begin{equation*}
  S_1 = \begin{pmatrix} 1.7 & 0 \\ 1 & 1 \end{pmatrix}, \qquad
  S_2 = (S_1 S_1^\top)^{1/2} \approx \begin{pmatrix} 1.59 & 0.59 \\ 0.59 & 1.29 \end{pmatrix}, \qquad
  S_3 = S_1 \begin{pmatrix} 1 & 0 \\ 0 & -1 \end{pmatrix} = \begin{pmatrix} 1.7 & 0 \\ 1 & -1 \end{pmatrix}.
\end{equation*}
All three satisfy $S_i S_i^\top = S_1 S_1^\top =: \Sigma$, so they induce the same interventional distributions $\nu_0 = \mu =  \normal(0, I_2)$ and $\nu_1 = \normal(0, \Sigma)$ and differ only in the conditional distributions $\nu_{1,v}$ (Figure~\ref{fig:toy-example-nu-a}(b--d)). In Setting~1, $S_1$ is the \ac{kr} map in ordering $(1,2)$, so $f \in \mathcal{C}_{(1,2)}^{\state{A}}$. In Setting~2, $S_2$ is the Brenier map, so $f \in \mathcal{C}_{\mathrm{ot}}^{\state{A}}$. In Setting~3, $S_3$ applies $S_1$ after reflecting the second coordinate of $U$. This leaves $\mu$ invariant but not $\mu_v$, and $f$ lies in neither class.

\paragraph{Conditional latent distributions.}
Let $\mathrm{s}_1, \dots, \mathrm{s}_K \subset \mathbb{R}^2$ denote the $K$ open angular sectors of
angle $2\pi/K$, centred at the angles $2\pi k/K$, $k = 0, \dots, K-1$, so that the boundary rays
lie at the angles $(2k-1)\pi/K$. For $K = 2$ the sectors are the right and left half-planes, and for
$K = 4$ they are the four quadrants rotated by $45^\circ$. Write
$\sigma \coloneq \sqrt{1 - \rho^2}$ and $a \coloneq \rho/\sigma$. Since $(W, U)$ is jointly Gaussian with
$\mathrm{Cov}(W, U) = \rho I_2$, we have $W \mid U = u \sim \mathcal{N}(\rho u, \sigma^2 I_2)$, and
$P(V = v) = 1/K$ by symmetry. Bayes' rule therefore gives
\begin{equation}\label{eq:toy-density}
  p_{U \mid V = v}(u) = K\, \varphi_2(u)\, \pi_v(u), \qquad
  \pi_v(u) \coloneq P\big(\rho u + \sigma Z \in \mathrm{s}_v\big), \quad Z \sim \mathcal{N}(0, I_2),
\end{equation}
where $\varphi_2$ is the standard normal density on $\mathbb{R}^2$: $\mu_v$ is
$\mathcal{N}(0, I_2)$ reweighted by the posterior probability that $W$ lies in sector $v$ given
$U = u$. Since $f_0 = \mathrm{id}$, $G_0 = \{S \in \mathrm{Diff}^1(\mathbb{R}^2) :
S_\# \mu_v = \mu_v \text{ for all } v\}$, and because the densities in \eqref{eq:toy-density} are
continuous and positive, $S \in G_0$ if and only if
\begin{equation}\label{eq:toy-cov}
  p_{U \mid V = v}(S(u))\, |\det J_S(u)| = p_{U \mid V = v}(u) \qquad \text{for all } u, v.
\end{equation}
Dividing \eqref{eq:toy-cov} for two strata shows that every $S \in G_0$ preserves the likelihood
ratios,
\begin{equation}\label{eq:toy-ratio}
  \frac{\pi_v(S(u))}{\pi_{v'}(S(u))} = \frac{\pi_v(u)}{\pi_{v'}(u)} \qquad \text{for all } u, v, v'.
\end{equation}

\paragraph{$K = 2$: the identified set has two points.}
With $\mathrm{s}_\pm = \{w : \pm w_1 > 0\}$ we have $\pi_\pm(u) = \Phi(\pm a u_1)$, so the ratio
$\pi_+/\pi_-$ is the function $t \mapsto \Phi(at)/\Phi(-at)$ of $u_1$ alone, which is strictly
increasing for $\rho > 0$. Hence \eqref{eq:toy-ratio} forces $S(u)_1 = u_1$, i.e.\
$S(u) = (u_1, m(u_1, u_2))$ for a $C^1$ function $m$ with $\partial_2 m \neq 0$. Inserting this
into \eqref{eq:toy-cov} and cancelling the common factor $2\varphi(u_1)\Phi(\pm a u_1)$ leaves
\[
  \varphi\big(m(u_1, u_2)\big)\,|\partial_2 m(u_1, u_2)| = \varphi(u_2) \qquad \text{for all } u.
\]
For fixed $u_1$ the map $t \mapsto m(u_1, t)$ is a $C^1$ bijection of $\mathbb{R}$ with
non-vanishing derivative, hence strictly monotone, and it transports $\mathcal{N}(0,1)$ to itself,
so it equals $t \mapsto t$ or $t \mapsto -t$. By continuity in $u_1$ the same alternative holds
for every $u_1$. Both alternatives do lie in $G_0$, and therefore
\[
  G_0 = \{\mathrm{id}, R\}, \qquad R \coloneq \operatorname{diag}(1, -1), \qquad
  T^f_{0\to 1}(G_0 \cdot y) = \{S_i y,\; S_i R y\}.
\]
The identified set consists of the true counterfactual and its image under the reflection of the
second latent coordinate. Since $S_3 = S_1 R$, Settings 1 and 3 induce the same family
$\{\nu_{a,v}\}_{a,v}$ for $K = 2$ and cannot be distinguished by any test.

\paragraph{$K = 4$: point identification.}
Let $Q$ denote the rotation by $45^\circ$. Since $\varphi_2$ and the conditional law of $W$ given
$U$ are invariant under rotations, $S \in G_0$ if and only if $\tilde S \coloneq Q S Q^{-1}$
satisfies \eqref{eq:toy-cov} and \eqref{eq:toy-ratio} with the sectors $Q\mathrm{s}_v$, which are
the quadrants $\mathrm{s}_{(s,t)} = \{w : s w_1 > 0,\ t w_2 > 0\}$, $s,t \in \{+,-\}$. For the
quadrants the posterior factorises, $\pi_{(s,t)}(u) = \Phi(s a u_1)\,\Phi(t a u_2)$. Applying
\eqref{eq:toy-ratio} to the strata $(+,+)$ and $(-,+)$ gives
$\Phi(a \tilde S(u)_1)/\Phi(-a \tilde S(u)_1) = \Phi(a u_1)/\Phi(-a u_1)$, hence
$\tilde S(u)_1 = u_1$ by strict monotonicity as in the case $K = 2$. Applying it to $(+,+)$ and
$(+,-)$ gives $\tilde S(u)_2 = u_2$ in the same way. Hence $\tilde S = \mathrm{id}$, so
$G_0 = \{\mathrm{id}\}$ and $T^f_{0\to1}(G_0 \cdot y) = \{S_i y\}$.

\paragraph{Additional figures.}
Figure~\ref{fig:toy-example-nu-a} shows $n = 3000$ samples with $K = 4$ and $\rho = 0.8$, and
Figure~\ref{fig:toy-example-identified-set} shows $n = 3000$ samples with $\rho = 0.9$, drawn from
the same latent sample for every $K$, so that only $V$ changes between the panels. The three
factual points are the sample points closest to three fixed locations. Figure~\ref{fig:toy-example-power-rho}(b)
shows $n = 2000$ samples with $K = 4$.

\paragraph{Test statistic and simulation settings.}
For the power experiment we use $K = 4$, sample sizes $n \in \{500, 2000, 8000\}$, and
$\rho \in \{0, 0.1, 0.2, 0.3, 0.45, 0.6, 0.8\}$. Since $\nu_0 = \mathcal{N}(0, I_2)$ and
$\nu_1 = \mathcal{N}(0, \Sigma)$ are known, we use the exact canonical outcome functions rather
than estimates: $h^{\mathcal{C}}_0 = \mathrm{id}$ and $h^{\mathcal{C}}_1 = S_1$ for
$\mathcal{C}_{(1,2)}$ and $h^{\mathcal{C}}_1 = S_2 = \Sigma^{1/2}$
for $\mathcal{C}_{\mathrm{ot}}$. These are the same in all three settings because
$S_i S_i^\top = \Sigma$ for $i = 1,2,3$. The power study thus isolates the effect of heterogeneity
on the test from estimation error in $h^{\mathcal{C}}$. The ACTG~175 example in
Section~\ref{sec:actg} uses estimated maps. For each run we draw $n$ i.i.d.\ samples of
$(V, A, Y)$, compute within each stratum $v$ James' statistic for the two samples
$\{(h^{\mathcal{C}}_0)^{-1}(Y_j) : A_j = 0, V_j = v\}$ and
$\{(h^{\mathcal{C}}_1)^{-1}(Y_j) : A_j = 1, V_j = v\}$, and sum over the $K$ strata. James'
statistic is the two-sample Hotelling statistic with unpooled covariance matrices,
$d_v^\top \hat\Sigma_v^{-1} d_v$ with $d_v$ the difference of the two sample means and
$\hat\Sigma_v$ the sum of the two estimated covariance matrices of the means. Mean independence is
a necessary condition for the conditional independence to hold but weaker than a full \ac{ci}
test. Under the null each statistic is asymptotically $\chi^2_2$, so the sum is compared to the
$\chi^2_{2K}$ quantile at level $\alpha = 0.05$. Rejection rates in
Figure~\ref{fig:toy-example-power-rho} are averaged over $500$ independent runs. Error bars show
$\pm$ one standard error of the estimated rejection rate. The green curves confirm that the test is
well calibrated.

\subsection{Real-world example: ACTG~175}\label{app:actg-details}

\begin{figure}[ht]
    \centering
    \includegraphics[width=0.75\linewidth]{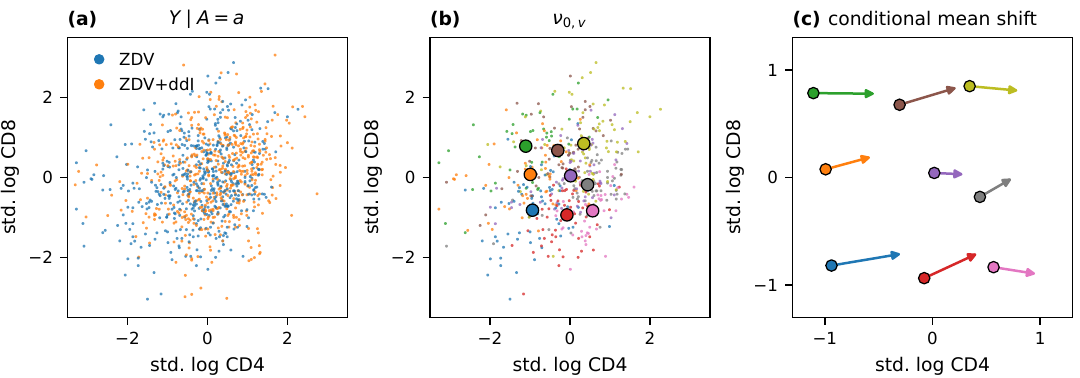}
    \caption{ACTG~175 outcomes. (a)~Standardised log CD4 and CD8 counts at week~20 by arm. (b)~The ZDV arm coloured by $V$. Circles mark the conditional means $\E[Y \mid A=0, V=v]$. (c)~Shift of the conditional means from ZDV ($A=0$) to ZDV+ddI ($A=1$).}
    \label{fig:actg-data}
\end{figure}

\paragraph{Data and preprocessing.}
We use the patients of the ZDV arm ($n_0 = 532$) and the ZDV+ddI arm ($n_1 = 522$) of the ACTG~175
data with CD4 and CD8 counts recorded at week~20. The outcome is the natural logarithm of the two
counts at week~20, standardised with one affine map for both arms (mean and standard deviation of
the pooled sample). The pooled standardisation matters for $\mathcal{C}_{\mathrm{ot}}$, whose
canonical map is not invariant to arm-specific rescaling. The covariate $V$ is the pair of
tercile indices of the baseline CD4 and CD8 counts, with terciles computed on the pooled sample,
so $V$ takes nine values with $30$ to $89$ patients per arm and stratum.
Figure~\ref{fig:actg-data} shows the outcomes and strata.

\begin{figure}[ht]
    \centering
    \includegraphics[width=0.9\linewidth]{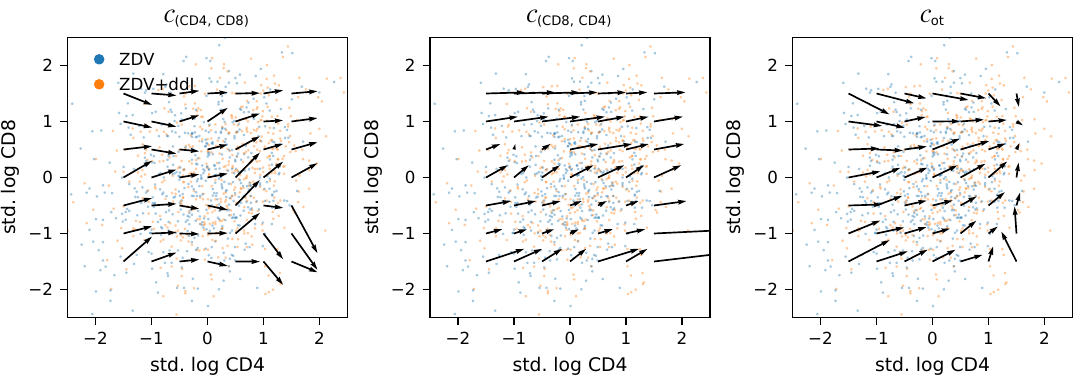}
    \caption{Estimated counterfactual maps $T^{h^{\mathcal{C}}}_{0\to1}$ from ZDV to ZDV+ddI for the three hypothesised model classes, drawn as arrows on a grid over the ZDV sample.}
    \label{fig:actg-cf-maps}
\end{figure}

\paragraph{Canonical outcome functions.}
For $\mathcal{C}_\pi$ we fit one autoregressive rational-quadratic spline flow per arm with base
distribution $\eta = \mathcal{N}(0, I_2)$ in the orderings (CD4, CD8) and (CD8, CD4). Each flow has
six spline bins on $[-3.5, 3.5]$ and an autoregressive conditioner with one hidden layer of $16$
units, and is trained by full-batch maximum likelihood for $2000$ steps with Adam (learning rate
$3 \cdot 10^{-3}$, weight decay $10^{-3}$, cosine learning-rate schedule) at fixed seeds. For
$\mathcal{C}_{\mathrm{ot}}$ we draw a reference sample of size $n_a$ from $\eta$ as a scrambled
Halton sequence mapped through the normal quantile function and exactly centred and whitened, solve
the assignment problem with quadratic cost between the reference sample and the observed outcomes
of arm $a$ exactly, and extend the resulting discrete map and its inverse to $\mathbb{R}^2$ by
$k$-nearest-neighbour regression on the matched pairs ($k = 15$). The maps in
Figure~\ref{fig:actg-cf-maps} are fitted on all patients of each arm.

\begin{figure}[t]
    \centering
    \includegraphics[width=0.45\linewidth]{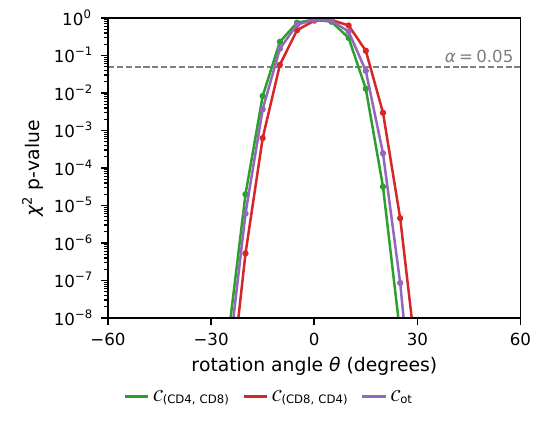}
    \caption{$p$-value of the \ac{ci} test for the hypotheses $h^\theta$, which transport $\eta$ to $\nu_1$ for every $\theta$ but rotate the treated latents by $\theta$. Outside the shown window $p < 10^{-8}$ for all three classes.}
    \label{fig:actg-rotation-sweep}
\end{figure}

\paragraph{Test details.}
We test $(h^{\mathcal{C}}_A)^{-1}(Y) \indep A \mid V$ with the statistic of
Appendix~\ref{app:details-toy-example}: James' statistic in each of the nine strata, summed, with
$\chi^2_{18}$ as the asymptotic null distribution. We also compute a permutation $p$-value by
permuting $A$ within strata ($1000$ permutations). The two agree to two decimals, and we report the
$\chi^2$ $p$-values in the main text and in Figure~\ref{fig:actg-rotation-sweep}. For the
triangular classes the latents entering the test are cross-fitted: the patients are split at random
into two halves, a flow per arm is fitted on each half and pulls back the patients of the other
half, so no patient is evaluated with a flow that was fitted on them. For
$\mathcal{C}_{\mathrm{ot}}$ the latent of a patient is the reference point they are assigned to,
which is an in-sample quantity. Figure~\ref{fig:actg-pulled-back-strata} shows the stratum means of
these latents.

\paragraph{Power on real data.}
For the reflection $R = \operatorname{diag}(1,-1)$ the test rejects for all three classes with
$\chi^2$ $p$-values between $10^{-170}$ and $10^{-143}$ and a permutation $p$-value of $0.001$, the
smallest value attainable with $1000$ permutations. For the rotation sweep we evaluate $R_\theta$ on
the grid $\theta \in \{-175^\circ, -170^\circ, \dots, 180^\circ\}$ with the same cross-fitted or
in-sample latents, rotating the latents of the ZDV+ddI arm by $\theta$.
Figure~\ref{fig:actg-rotation-sweep} shows the $\chi^2$ $p$-values. The rotation angles quoted in
Section~\ref{sec:actg} are computed from the maps fitted on all patients of each arm: for every
patient we take the signed angle, counter-clockwise positive, from their pulled-back latent under
one class to their pulled-back latent under another, and report the median over all patients of
both arms (from $\mathcal{C}_{(\mathrm{CD4},\mathrm{CD8})}$ to
$\mathcal{C}_{(\mathrm{CD8},\mathrm{CD4})}$: $13.6^\circ$ and from either \ac{kr} class to
$\mathcal{C}_{\mathrm{ot}}$: $7.1^\circ$ and $-6.7^\circ$) and separately over the patients of each
arm ($11.3^\circ$ and $17.8^\circ$, $5.4^\circ$ and $9.2^\circ$, $-5.4^\circ$ and $-8.0^\circ$ for
ZDV and ZDV+ddI, respectively). The difference between the two arms is what the \ac{ci} test can
detect.

\paragraph{Additional figures.}

\end{document}